%% file: main.tex
\documentclass[sigconf]{acmart}
\usepackage{booktabs} % For formal tables
\usepackage{tikz-cd} 
\usepackage[utf8]{inputenc}
\usepackage{adjustbox}
\usepackage{multirow}
\usepackage{multicol}
\usepackage[]{algorithm2e}
\usepackage{arydshln}
\usepackage{pgfplots}
\usepackage{flushend}
\usetikzlibrary{patterns}
\usepackage{balance}
\usepackage{subfigure}
\pgfplotsset{compat=1.11,
        /pgfplots/ybar legend/.style={
        /pgfplots/legend image code/.code={%
        \draw[##1,/tikz/.cd,bar width=3pt,yshift=-0.2em,bar shift=0pt]
                plot coordinates {(0cm,0.8em)};},
},
}
\input{parts/macros.tex}

\def\BibTeX{{\rm B\kern-.05em{\sc i\kern-.025em b}\kern-.08emT\kern-.1667em\lower.7ex\hbox{E}\kern-.125emX}}

\copyrightyear{2019}
\acmYear{2019}
\setcopyright{rightsretained}
\acmConference[CIKM '19]{CIKM conference}{Nov 03--07, 2019}{Beijing, China}

\author{Julien Romero}
\affiliation{Télécom Paris}
\email{romero@telecom-paris.fr}
\author{Simon Razniewski}
\affiliation{Max Planck Institute for Informatics}
\email{srazniew@mpi-inf.mpg.de}
\author{Koninika Pal}
\affiliation{Max Planck Institute for Informatics}
\email{kpal@mpi-inf.mpg.de}
\author{Jeff Z. Pan}
\affiliation{University of Aberdeen}
\email{jeff.z.pan@abdn.ac.uk}
\author{Archit Sakhadeo}
\affiliation{Max Planck Institute for Informatics}
\email{architsakhadeo@gmail.com}
\author{Gerhard Weikum}
\affiliation{Max Planck Institute for Informatics}
\email{weikum@mpi-inf.mpg.de}

\renewcommand{\shortauthors}{J. Romero et al.}

\setcopyright{acmlicensed}
\begin{document}

\fancyhead{}

%\title[Salient Commonsense Properties from Query Logs and Answer Snippets]{Salient Commonsense Properties\\ from Query Logs and Answer Snippets}
\title[Commonsense Properties from Query Logs and Question Answering Forums]{Commonsense Properties from Query Logs \texorpdfstring{\\ }{} and Question Answering Forums}

\begin{abstract}
%%%Motivation and Problem
Commonsense knowledge about object properties, human behavior and
general concepts 
is crucial for 
robust AI applications.
However, automatic acquisition of this knowledge is challenging
because of sparseness and bias in online sources.
This paper presents Quasimodo, 
a methodology and tool suite for distilling commonsense properties
from non-standard web sources.
We devise novel ways of tapping into
search-engine query logs and 
QA forums,
and combining the
resulting candidate assertions with statistical cues from
encyclopedias, books and image tags in a 
corroboration
step.
Unlike prior work on commonsense knowledge bases, 
Quasimodo focuses on salient properties that are typically
associated with certain objects or concepts. 
Extensive evaluations, including extrinsic use-case studies, 
show that Quasimodo provides better
coverage than state-of-the-art baselines with comparable
quality.
\end{abstract}

\copyrightyear{2019}
\acmYear{2019}
\acmConference[CIKM '19]{The 28th ACM International Conference on Information and Knowledge Management}{November 3--7, 2019}{Beijing, China}
\acmBooktitle{The 28th ACM International Conference on Information and Knowledge Management (CIKM '19), November 3--7, 2019, Beijing, China}
\acmPrice{15.00}
\acmDOI{10.1145/3357384.3357955}
\acmISBN{978-1-4503-6976-3/19/11}

\maketitle

\input{parts/intro.tex}
\input{parts/relatedwork.tex}

\input{parts/system.tex}

\input{parts/querylogs.tex}

\input{parts/cleaningfusion.tex}
\input{parts/ranking.tex}
\input{parts/refinementgrouping.tex}

\input{parts/experiments.tex}
\input{parts/conclusion.tex}

%\newpage
\bibliographystyle{myplain}
\bibliography{refs}

\newpage

\appendix

\input{parts/appendix.tex}

\end{document}

%% file: parts/macros.tex
\newcommand{\sr}[1]{\textit{\textcolor{green}{SR: #1}}}

\newcommand{\squishlist}{
 \begin{list}{$\bullet$}
  { \setlength{\itemsep}{0pt}
     \setlength{\parsep}{3pt}
     \setlength{\topsep}{3pt}
     \setlength{\partopsep}{0pt}
     \setlength{\leftmargin}{1.5em}
     \setlength{\labelwidth}{1em}
     \setlength{\labelsep}{0.5em} } }

\newcommand{\squishend}{
    \end{list}  }      

\newcommand{\fact}[1]{{\small\tt #1}}

%% file: parts/intro.tex
\section{Introduction}

\subsection{Motivation and Goal}

%first par:
%CSKB old theme in AI
%about properties of everyday objects, typical human behavior and emotions, 
%and general plausibility invariants
%witness revival: crucial for robust and explainable AI
%use cases: language comprehension, chatbots, visual and multimodal content understanding, and more

Commonsense knowledge (CSK for short) 
%and reasoning 
is an old theme in AI, 
already envisioned by McCarthy in the 1960s~\cite{mccarthy1960programs} and later pursued
by AI pioneers like Feigenbaum~\cite{feigenbaum1984knowledge} and Lenat~\cite{lenat1995cyc}.
The goal is to equip machines with knowledge of properties of everyday objects
(e.g., bananas are yellow, edible and sweet), typical human behavior and emotions
(e.g., children like bananas, children learn at school, death causes sadness)
and general plausibility invariants (e.g., a classroom of children should also have a teacher).
%
%Common sense knowledge includes basic assertions like \emph{(monkeys, like, bananas)}, \emph{(rain, causes, %wet ground)}, or \emph{(climate change, threatens, biodiversity)}. 
In recent years, research on automatic acquisition of such knowledge has been revived,
driven by the pressing need for human-like AI systems with robust and explainable behavior. 
Important use cases of CSK include the interpretation of user intents in search-engine queries,
question answering, versatile chatbots, language comprehension, visual content understanding, and more.
%
%Such knowledge is a fundamental asset for applications such as question answering, dialogue systems, or 
%reasoning and planning. This holds especially in the face of recent trends towards large-scale end-to-end 
%learning, which yields impressive results on selected tasks, but greatly limits the ability to reuse and 
%generalize learned models.
%GW: commented out, as this sounds a bit too general - may not appeal to IR community

%2nd par:
%examples (give 2 to 4 compelling examples)
%1) QA over general texts: 
%2) chatbot Tay: holocaust jokes are tasteless and offend most people
%3) visual/multimodal content: 
%4) recommenders: for birthday party - children like sweet drinks, no alcohol

{\bf Examples:} 
A keyword query such as ``Jordan weather forecast'' is ambiguous, but CSK should tell
the search engine that this refers to the country
and not to a basketball player or machine learning professor.
A chatbot should know that racist jokes are considered tasteless and would offend its users;
so CSK could have avoided the 2016 PR disaster of the Tay chatbot.\footnote{\scriptsize\url{www.cnbc.com/2018/03/17/facebook-and-youtube-should-learn-from-microsoft-tay-racist-chatbot.html}}
In an image of a meeting at an IT company where one person wears a suit and another person is in
jeans and t-shirt, the former is likely a manager and the latter an engineer.
Last but not least, a ``deep fake'' video where Donald Trump rides on the back of a tiger
could be easily uncovered by knowing that tigers are wild and dangerous and, if at all,
only circus artists would do this.

%3rd par:
%goal of this work is ...
%consider properties that can be cast into SPO triples
%aim for typical properties, not possible ones

The goal of this paper is to advance the automatic acquisition of salient
commonsense properties from online content of the Internet.
For knowledge representation, we focus on simple assertions in the form of
subject-predicate-object (SPO) triples such as {\small\tt children like banana}
or {\small\tt classroom includes teacher}.
Complex assertions, such as Datalog clauses, and 
logical reasoning over these are outside our scope.

A major difficulty that prior work has struggled with is the sparseness and bias
of possible input sources.
Commonsense properties are so mundane that they are rarely expressed in explicit terms
(e.g., countries or regions have weather, people don't).
Therefore, typical sources for information extraction like Wikipedia are fairly useless for CSK.
Moreover, online contents, like social media (Twitter, Reddit, Quora etc.), fan communities (Wikia etc.)
and books or movies, are often heavily biased and do not reflect typical real-life situations.
For example, 
%previously constructed 
existing
CSK repositories contain odd triples such as
{\small\tt banana located\_in monkey's\_hand}, {\small\tt engineer has\_property conservative}, 
{\small\tt child make choice}.
%
%Collecting and structuring common-sense knowledge is difficult for several reasons: The nature of common-sense is that it is often implicitly known by %humans, but hardly expressed in structured or semi-structured form, thus making automated extraction difficult. Similarly, it is very difficult to %establish a notion of salience on common-sense, i.e., to determine what are relevant and/or interesting statements about a given subject.

%%%%%%%%%%%%%%%%%%%%%%%%%%%%%%%%%%%%%%%%%%%%

\subsection{State of the Art and Limitations}

%this subsection:
%biggest CSKBs: ConceptNet, WebChild, TupleKB
%each with strengths and major limitations
%ConceptNet: few predicates, low coverage of properties per S, 
%WebChild: very mixed precision, focus on possible properties
%TupleKB: low coverage, source bias

Popular knowledge bases like DBpedia, Wikidata or Yago
have a strong focus on encyclopedic knowledge about
individual entities like (prominent) people, places etc.,
and do not cover commonsense properties of general concepts.
The notable exception is the inclusion of 
SPO triples for the {\small\tt (sub-)type} 
(aka. {\small\tt isa}) predicate, for example, {\small\tt banana type fruit}. Such triples are ample especially in
Yago (derived from Wikipedia categories and imported from
WordNet). Our focus is on additional
properties beyond {\small\tt type}, which are absent
in all of the above knowledge bases.

The most notable projects on constructing commonsense knowledge bases are
Cyc \cite{lenat1995cyc}, ConceptNet \cite{conceptnet}, WebChild \cite{webchild}
and Mosaic TupleKB \cite{tuplekb}. Each of these has specific strengths and limitations.
The seminal Cyc project solely  relied on human experts for codifying logical assertions,
with inherent limitations in scope and scale.
ConceptNet used crowdsourcing for scalability and better coverage, but is limited to only
a few different predicates like {\small\tt has\_property}, {\small\tt located\_in}, {\small\tt used\_for}, {\small\tt capable\_of}, 
{\small\tt has\_part}
and {\small\tt type}. 
Moreover, the crowdsourced inputs often take noisy, verbose or uninformative forms 
(e.g., {\small\tt banana type bunch}, {\small\tt banana type herb}, {\small\tt banana has\_property good\_to\_eat}).
WebChild tapped into book n-grams and image tags to overcome the bias in many Web sources.
It has a wider variety of 20 predicates  and is much larger, but contains
a heavy tail of noisy and dubious triples -- due to its focus on possible properties rather than typical ones
(e.g., engineers are conservative, cool, qualified, hard, vital etc.).
TupleKB is built by carefully generating search-engine queries on specific domains and performing
various stages of information extraction and cleaning on the query results.
Despite its clustering-based cleaning steps, it contains substantial noise and is limited in scope
by the way the queries are formulated.

The work in this paper aims to overcome the bottlenecks of these prior projects while preserving their
positive characteristics. In particular, we aim to achieve high coverage, like WebChild, with high precision
(i.e., a fraction of valid triples), like ConceptNet. 
In addition, we strive to acquire properties for a wide range of predicates - more diverse and refined
than ConceptNet and WebChild, but without the noise that TupleKB has acquired.

%\jr{Talk about Wikidata, comparison} \sr{I put a paragraph into related work}
%%%GW: addressed above and resolved

%Several works have targeted the construction of large common-sense knowledge bases in the past. An early approach is the CYC project, which had 
%experts write down common-sense assertions and rules~\cite{lenat1995cyc}. Another prominent approach, ConceptNet, used crowdsourcing for better 
%scalability~\cite{conceptnet}. Since then, various projects have focused on automatic common-sense knowledge base (CSKB) construction from text and 
%image sources, most prominently WebChild~\cite{webchild} and TupleKB~\cite{tuplekb}.  

%These efforts have produced encouraging initial repositories of common-sense assertions. Nevertheless, they exhibit important limitations, in 
%particular w.r.t.\ their generality, relevance, size, and informativeness of predicates.

%%%%%%%%%%%%%%%%%%%%%%%%%%%%%%%%%%%%%%%%%%%%

\subsection{Approach and Challenges}

%desiderata:
%coverage like WebChild, precision like ConceptNet,
%refined predicates, focus on typical properties

% one par on approach
%approach in a nutshell
%unique points are:
%* tap into new sources for diversity: query logs and answer snippets
%* fuse evidence from multiple sources for corroboration
%* refine and group assertions by tri-factorization
%* ranking by saliency (typical properties)

This paper puts forward {\em Quasimodo}\footnote{The name stands for: Query Logs and 
%Answer Snippets 
QA Forums
for Salient Commonsense Definitions.
Quasimodo is the main character in Victor Hugo's novel ``The Hunchback of Notre Dame'' who epitomizes
human preconception and also exhibits unexpected traits.},
 a framework and tool for scalable automatic acquisition of commonsense properties.
Quasimodo is designed to tap into 
%multiple sources of noisy cues and combines them in a knowledge fusion step for
%corroboration. 
non-standard sources where questions rather than statements
provide cues about commonsense properties.
This leads to noisy candidates for populating a
commonsense knowledge base (CSKB).
To eliminate false positives, we have devised a subsequent
cleaning stage, where corroboration signals are obtained 
from a variety of sources and combined by learning a regression model.
This way, 
Quasimodo reconciles wide coverage with high precision.
In doing this, it focuses on salient properties which typically occur for common concepts, while eliminating possible but atypical
and uninformative output. This counters the reporting bias - frequent mentioning of sensational but unusual and unrealistic properties
(e.g., pink elephants in Walt Disney's Dumbo).

%Our framework utilizes targeted novel sources for information extraction, affirmative questions from search engine query logs and discussion and 
%question-answering forums. In a recall-oriented extraction stage, it utilizes customized open information extraction to extract candidate triples. 
%These triples are normalized in several carefully crafted steps, then fed into a precision-oriented validation stage, where frequency information 
%across sources, along with linguistic cues and frequencies in search engine snippets are used to estimate typicality and correctness.

The new sources that we tap into 
for gathering candidate assertions
are search-engine query logs 
and question answering forums like Reddit, Quora etc.
%, and answer snippets from search engines for corroboration.
%These are unavailable outside of the big industrial labs, but can be sampled by using
%search-engine interfaces in a creative way.
%As for query logs, Quasimodo generates queries in a judicious way 
%and collects
%auto-completion suggestions.
Query logs are unavailable outside industrial labs,
but can be sampled by using search-engine interfaces
in a creative way.
To this end, Quasimodo 
generates queries in a judicious way and collects auto-completion suggestions.
%As for answer snippets, we devise a self-training (i.e., unsupervised) way of
%extracting candidate assertions from the ungrammatical text of top-100 result snippets --
%for a carefully selected set of probing queries 
%and under a tight budget limit for daily Google queries.
The subsequent corroboration stage
%fusion of candidates 
%and statistical evidence
%from different sources 
harnesses 
%additional 
statistics from search-engine answer snippets,
Wikipedia editions, 
Google Books and image tags by means of a learned regression model.
This step is geared to eliminate noisy, atypical, and uninformative properties.

A subsequent ranking step further enhances the knowledge quality in terms
of typicality and saliency.
Finally, to counter noisy language diversity, reduce semantic redundancy, and
canonicalize the resulting commonsense triples to a large extent, 
Quasimodo includes a novel way of clustering the triples that result from the fusion step.
This is based on a tri-factorization model for matrix decomposition.

%The superiority of this pipeline w.r.t.\ ConceptNet, WebChild and TupleKB is evaluated along a set of dimensions, covering the intrinsic metrics of 
%size, correctness, typicality and ranking ability, and extrinsic evaluations in textbook coverage, question answering and Taboo forbidden term 
%prediction.
%GW: this fits better under Contributions

% one par on challenges:
%challenges to be overcome:
%? cope with bias in query log
%? cope with ungrammatical text from answer snippets
%? lack of (explicit) training data (for supervised learning)
%? noisy output with semantic redundancy

Our approach faces two major challenges:
\squishlist
\item coping with the heavy reporting bias in cues from query logs,
potentially leading to atypical and odd properties,
%\item coping with ungrammatical text from answer snippets which
%render standard NLP techniques useless (e.g., dependency parsing in Open IE), 
%\item coping with the lack of explicit training data which makes 
%supervised learning inappropriate (e.g., deep neural networks), and
\item coping with the noise, language diversity, and semantic redundancy
in the output of information extraction methods.
\squishend

The paper shows how these challenges can be (largely) overcome.
Experiments demonstrate the practical viability of Quasimodo
and its improvements over prior works.

%%%%%%%%%%%%%%%%%%%%%%%%%%%%%%%%%%%%%%%%%%%%

\subsection{Contributions}

% crisp list with
%* novel ways of tapping query logs and answer snippets for knowlege acquisition
%* novel way of refining and organizing commonsense knowledge
%* high-coverage commonsense KB with salient properties on two topical domains (professions and animals)
%* systematic evaluation of KB quality and benefit in extrinsic use cases

The paper makes the following original contributions:
%\begin{enumerate}
\squishlist
     \item a complete methodology and tool for multi-source acquisition of typical and salient commonsense properties
%methodology, along with a sample instantiation that yields knowledge of significantly higher coverage and precision than existing sources; 
      with  principled methods for 
      %knowledge fusion and cleaning,
      corroboration, ranking and refined grouping,
    \item novel ways of tapping into non-standard input sources like query logs
    and QA forums,
    %and answer snippets
%    \item The exploitation of affirmative question sources, allowing to extract knowledge that would hardly be present explicitly anywhere else;
    \item a high-quality knowledge base of ca. 2.26 million salient properties for ca. 80,000 concepts, 
    %on the domains of animals and human occupations, 
       publicly available as a research resource,
      \footnote{\url{https://www.mpi-inf.mpg.de/departments/databases-and-information-systems/research/yago-naga/commonsense/quasimodo/}}
      %\kp{isn't it 1.23 million facts ??}
%\sr{Needs rethinking how to word it: essentially there will be two resources: a) Highest quality KB on two small domains, 50 animals and 50 professions, b) Snippet-less KB for 13,000 concepts.}
%
%which outperforms state-of-the-art CSKBs by at least ZYX\% in intrinsic preference and YXZ\% in coverage, 
%and YXZ\% in extrinsic use cases of science question answering and ZXY\% for summarization.
     \item an experimental evaluation and comparison to ConceptNet, WebChild, and TupleKB which shows major gains in coverage and quality, and
     \item experiments on extrinsic tasks like language games (Taboo word guessing) and question answering.
\squishend
Our code will be made available on Github\footnote{\url{https://github.com/Aunsiels/CSK}}.

%% file: parts/relatedwork.tex
\section{Related Work}

\noindent{\bf Commonsense Knowledge Bases (CSKBs).}
%Common sense is considered a major component of intelligent applications and machines, and has recently been brought to central attention in AI research.\footnote{\url{https://www.geekwire.com/2018/paul-allen-invest-125m-new-common-sense-ai-project}} 
The most notable projects on building large commonsense knowledge bases
are the following.

\vspace{-5pt}
\paragraph{Cyc:}
The Cyc project
%, started in 1984 by Douglas Lenat, represents 
was the first major effort towards collecting and formalizing general world knowledge~\cite{lenat1995cyc}. 
%In the project, 
Knowledge engineers manually compiled knowledge, in the form of 
grounded assertions and logical rules. 
%Arguably, the scope of the project was underestimated 
%and the efforts were later largely discontinued. 
Parts of Cyc were released to the public as OpenCyc in 2002,
but these parts mostly focus on concept taxonomies, that is,
the {\small\tt (sub-)type} predicate.

\vspace{-5pt}
\paragraph{ConceptNet:}
%In the Open Mind Common Sense project, 
Crowdsourcing has been used to construct ConceptNet, a triple-based semantic network of commonsense assertions about general objects~\cite{DBLP:conf/lrec/SpeerH12,conceptnet}. ConceptNet contains 
%1.334.425 assertions for  842.532 subjects 
ca. 1.3 million assertions for ca. 850,000 subjects
(counting only English assertions and semantic relations, i.e., discounting relations like \fact{synonym} or \fact{derivedFrom}). 
The focus is on a small number of broad-coverage predicates, namely,
{\small\tt type, \!\! locationOf, usedFor, capableOf, hasPart}.
ConceptNet is one of the highest-quality and most widely used CSK resources.

\vspace{-5pt}
\paragraph{WebChild:}
WebChild has been automatically constructed from book n-grams (and, to a smaller degree, image tags) by a pipeline of information extraction,
statistical learning and constraint reasoning methods \cite{DBLP:conf/wsdm/TandonMSW14,webchild}. 
WebChild contains ca. 13 million assertions, 
and covers 20 distinct predicates such as
{\small\tt hasSize, hasShape, physicalPartOf, memberOf}, etc.
It is the biggest of the publicly available commonsense knowledge bases,
with the largest slice being on part-whole knowledge \cite{DBLP:conf/aaai/TandonHURRW16}.
However, a large mass of WebChild's contents is in the long tail of
possible but not necessarily typical and salient properties.
So it comes with a substantial amount of noise
and non-salient contents.

%For instance, it makes heavy use of subclass relationships in order to propagate the fact \fact{human has cheekbone} to \fact{opera\_singer has cheekbone}.

% Figure from system overview section, placed here so it appears on the same page as that section (hack)
\begin{figure}
\includegraphics[scale=0.35]{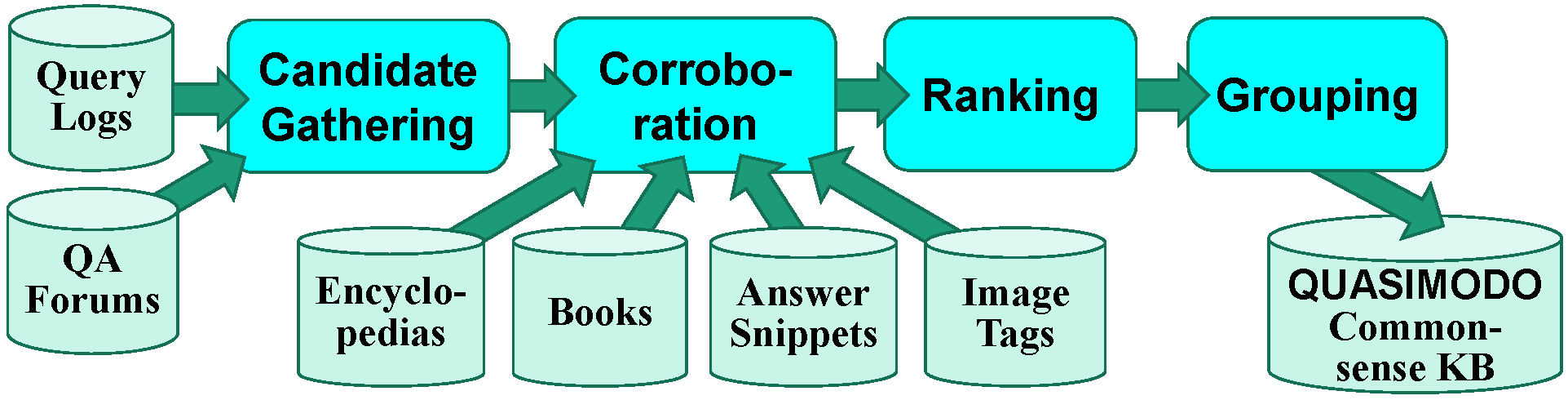}\caption{Quasimodo system overview.}
\label{fig:overview}
\vspace{-0.4cm}
\end{figure}

\vspace{-5pt}
\paragraph{Mosaic TupleKB:}
The Mosaic project at AI2 aims to collect commonsense knowledge in various forms, from grounded triples to procedural knowledge
with first-order logic.
TupleKB, released as part of this ongoing project,
is a collection of triples for the science domain, compiled 
by generating domain-specific queries and extracting assertions
from the resulting web pages. 
A subsequent cleaning step, based on integer linear programming,
clusters triples into groups.
TupleKB contains 
%282.595 triples for 28.079 subjects.
ca. 280,000 triples for ca. 30,000 subjects.

\vspace{-5pt}
\paragraph{Wikidata:} This collaboratively built knowledge base
is mostly geared to organize encyclopedic facts about
individual entities like people, places, organizations etc. \cite{vrandevcic2014wikidata,DBLP:conf/semweb/MalyshevKGGB18}.
It contains more than 400 million assertions for more than 50 million items.
This includes some world knowledge about general concepts, like {\small\tt type}
triples, but this coverage is very limited. For instance, 
Wikidata neither knows that birds can fly nor that elephants have trunks.

\vspace{-5pt}
\paragraph{Ascent:} The Ascent project~\cite{ascent} provides a recent extension of the Quasimodo approach. Its focus is on precision, and more refined knowledge representation, yet it relies on general websites retrieved via a search engine, and thus is not easily scalable.

%%%%%%%%%%%%%%%%%%%%%%%%%%%%%%%%%%%%%%%%%%%%%%%%%%%%%%%%%%%%

\vspace*{0.2cm}
\noindent{\bf Use Cases of CSK.}
%\paragraph{Usage of common sense}
Commonsense knowledge and reasoning are  instrumental in 
a variety of applications in natural language processing,
computer vision, and AI in general.
These include question answering, especially for 
general world comprehension~\cite{swag} and science questions~\cite{schoenick2016moving}. 
Sometimes, these use cases also involve additional
reasoning (e.g., \cite{DBLP:conf/emnlp/TandonDGYBC18}), where CSK contributes, too.
Another NLP application is dialog systems and chatbots
(e.g., \cite{young2018augmenting}),
where CSK adds plausibility priors to language
generation.
%It is also used in dialogue systems, with the aim to make system responses appear more targeted and natural~\cite{young2018augmenting}.
%

%\paragraph{Spatial Commonsense}
%Commonsense is also important in visual contexts. Besides WebChild, various other works explore the acquisition of common sense knowledge from image tags and spatial collocation information~\cite{xu2018automatic,yatskar2016stating}.
For visual content understanding, such as object detection
or caption generation for images and videos,
CSK can contribute as an informed prior
 about spatial co-location derived, for example, 
from image tags, and about human activities and 
associated emotions
(e.g., \cite{xu2018automatic,yatskar2016stating,DBLP:conf/wsdm/ChowdhuryTFW18}).
In such settings, CSK is an additional input to
supervised deep-learning methods.

%%%%%%%%%%%%%%%%%%%%%%%%%%%%%%%%%%%%%%%%%%%%%%%%%%%%%%%%%%%%

\vspace*{0.2cm}
\noindent{\bf Information Extraction from Query Logs.}
%\paragraph{Information extraction from query logs}
%Query logs are known to be a powerful information source. 
Prior works have tapped into query logs for 
goals like query recommendation
(e.g., \cite{cao2008context})
and extracting semantic relationships between search terms,
like synonymy and hypernymy/hyponymy 
(e.g., \cite{baeza-yates2007,DBLP:conf/sigmod/WuLWZ12,DBLP:conf/emnlp/Pasca13,DBLP:conf/acl/PascaD08}).
The latter can be seen as gathering triples for
CSK, but its sole focus is on the {\small\tt (sub-)type}
predicate -- so the coverage of the predicate space
is restricted to class/type taxonomies.
Moreover, these projects were carried out on 
full query logs within industrial labs of 
search-engine companies.
In contrast, Quasimodo addresses a much wider space
of predicates and operates with an original way of
sampling query-log-derived signals via 
auto-completion suggestions.
To the best of our knowledge, no prior work has
aimed to harness auto-completion for CSK acquisition
(cf. \cite{autocomplete}).
%
%With Quasimodo, we specifically make use of autocomplete functionalities, a feature now common in most major search engines and custom retrieval frameworks~\cite{autocomplete}.

The methodologically closest work to ours
is \cite{DBLP:conf/cikm/Pasca15}. 
Like us, that work used interrogative patterns
(e.g. ``Why do \dots'') to mine query logs
-- with full access to the search-engine company's logs.
Unlike us, subjects, typically classes/types such as
``cars'' or ``actors'', were merely associated with
salient phrases from the log rather than 
extracting complete triples. One can think of this
as organizing CSK in SP pairs where P is a textual phrase
that comprises both predicate and object but cannot
separate these two.
Moreover, \cite{DBLP:conf/cikm/Pasca15} restricted itself
to the extraction stage and used simple scoring
from query frequencies, whereas we go further by
leveraging multi-source signals in the corroboration stage
and refining the SPO assertions into semantic groups.

%% file: parts/system.tex
%\section{Design Rationale and System Overview}
\section{System Overview}

%Given the challenges in 1) retrieving ``obvious'' common-sense knowledge, and 2) separating typical and salient assertions from obscure and/or sensational statements, our rationale in designing Quasimodo is to 
Quasimodo is designed to cope with the high noise and potentially strong bias
in online contents. It taps into query logs via auto-completion suggestions
as a non-standard input source. However, frequent queries -- which are the
ones that are visible through auto-completion -- are often about sensational
and untypical issues. 
%A typical example would be ``Why are programmers \dots''
%completed into ``\dots so rude?''.
%A typical example would be ``Why are hockey players \dots''
%completed into ``\dots so hot?''.
Therefore, Quasimodo combine a recall-oriented candidate gathering phase with two subsequent phases for cleaning, refining, and ranking assertions.   
Figure~\ref{fig:overview} gives a pictorial overview of the system architecture.

\noindent{\bf Candidate Gathering.}
In this phase, we  
extract candidate triples from some of the world's largest
sources of the ``wisdom of crowds'', namely, search-engine query logs
and question answering forums such as Reddit or Quora.
While the latter can be directly accessed via search APIs,
query logs are unavailable outside of industrial labs.
Therefore, we creatively probe and sample
this guarded resource by means of generating queries and observing 
auto-completion suggestions by the search engine. The
resulting suggestions
are typically among the statistically frequent queries.
%\textbf{two of the world's best data sources of human knowledge}, web search query autocompletion and web search result snippets. 
%
%Query autocompletion is representative for the largest collection of observations that people wonder about. Search result snippets allow readily-ranked access to the world's largest text corpora, as stored in search engine indexes. Our access to these sources is carefully crafted - text corpora of search engines can only meaningfully be accessed with targeted queries, so we access them with subject-object pairs (SO). Autocompletion, on the other hand, works only for very short input sequences, thus we access it by patterns crafted around subjects (S) alone.
As auto-completion works only for short inputs of a few words, 
we generate queries that are centered on candidate subjects, the S argument
in the SPO triples that we aim to harvest.
%%%GW: additional details, especially the focus on interrogative queries
%%%rather than assertive ones, will be given in Section 4
Technical details are given in Section \ref{sec:gathering}.

%\noindent{\bf Cleaning and Fusion.} 
\noindent{\bf Corroboration.} 
This phase is precision-oriented, aiming to eliminate false positives from the
candidate gathering.
We consider candidates as invalid for three possible reasons:
1) they do not make sense (e.g., {programmers eat python});
2) they are not typical properties for the instances of the S concept
(e.g., {programmers drink espresso});
3) they are not salient in the sense that they are immediately
associated with the S concept by most humans
(e.g., {programmers visit restaurants}).
To statistically check to which degree these aspects are satisfied,
Quasimodo harnesses corroboration signals in a multi-source 
%fusion
scoring
step.
This includes standard sources like Wikipedia articles and books,
which were used in prior works already,
but also non-standard sources like image tags and
answer snippets from search-engine queries.
%\textit{cleaning and fusion phase}, we use crafted rules to filter and normalize extractions, and build a supervised classification model using features such as Wikipedia and image co-occurrence to filter triples that do not make sense at all.
Technical details are given in Section \ref{sec:cleaning}.

\noindent{\bf Ranking.} To identify typical and salient triples,
we devised a probabilistic ranking model with the corroboration scores as input signal.
This stage is described in Section \ref{sec:ranking}.

%\noindent{\bf Refinement and Grouping.}
\noindent{\bf Grouping.}
For this phase, we have devised a clustering 
method based on the model of tri-factorization for matrix decomposition \cite{DBLP:conf/kdd/DingLPP06}.
%in order to detect semantically equivalent predicate phrases, to refine generic %predicates, and to cautiously predict new assertions.
The output consists of groups of SO pairs and P phrases linked to each other.
So we semantically organize and refine both the concept arguments (S and O)
in a commonsense triple and the way the predicate (P) is expressed in language.
Ideally, this would canonicalize all three components, in analogy to
what prior works have achieved for entity-centric encyclopedic knowledge bases
(e.g., \cite{DBLP:conf/www/SuchanekSW09,DBLP:conf/cikm/GalarragaHMS14}).
However, commonsense assertions are rarely as crisp as facts about 
individual entities, and often carry subtle variation and linguistic diversity
(e.g., {\small\tt live in} and {\small\tt roam in} for animals being near-synonymous
but not quite the same).
Our clustering method also brings out refinements of predicates.
This is in contrast to prior work on CSK which has mostly restricted itself
to a small number of coarse-grained predicates like {\small\tt partOf, usedFor, locatedAt}, etc.
Technical details are given in Section \ref{sec:grouping}.

%% file: parts/querylogs.tex
%\section{Candidate Gathering from Query Logs and Forums}
\section{Candidate Gathering}
\label{sec:gathering}

The key idea for this phase is to utilize questions as a
source of human commonsense.
For example, the question \emph{``Why do dogs bark?''} implicitly conveys the user's knowledge that dogs bark.
Questions of this kind are posed in QA forums,
such as Reddit or Quora, but their frequency and
coverage in these sources alone is not sufficient
for building a comprehensive knowledge base.
Therefore, we additionally tap into query logs from
search engines, sampled through observing 
auto-completion suggestions.
Although most queries merely consist of 
a few keywords, there is a substantial fraction of
user requests in interrogative form \cite{DBLP:conf/www/WhiteRY15}.
%As question logs of search engines are not directly accessible, we creatively probe them via autocompletion. Also, we generically transform questions into assertional sentences, in order to be able to extract facts using standard OpenIE techniques. Finally, we apply a set of normalization operations in order to obtain better interpretable common-sense assertions.

\subsection{Data Sources}

Quasimodo exploits two data sources: (i) QA forums, which return
questions in user posts through their search APIs, and (ii) query logs from major search engines, which are sampled by
generating query prefixes and observing their auto-completions.

\vspace*{0.1cm}
\noindent{\bf QA forums.}
We use four different QA forums:
Quora, 
%(Alexa rank 91) 
%is presumably the most well known general-purpose QA forum.
Yahoo! Answers\footnote{\url{https://answers.yahoo.com} and \url{https://webscope.sandbox.yahoo.com}},
%is also very well known and gives access to part of the questions for research purposes.
Answers.com,
%(Alexa rank 2,500) 
%is a similarly themed site, and although less popular, allows easier data access. 
and
Reddit.
%(Alexa rank 18) 
%is a very popular discussion forum, for which data dumps are available. In total, we collected 6,529,853 questions from QA forums.
The first three are online communities for general-purpose QA
across many topics,
and Reddit is a large discussion forum with a wide variety of topical categories.

\begin{figure}
    \centering
    \includegraphics[width=0.2\textwidth]{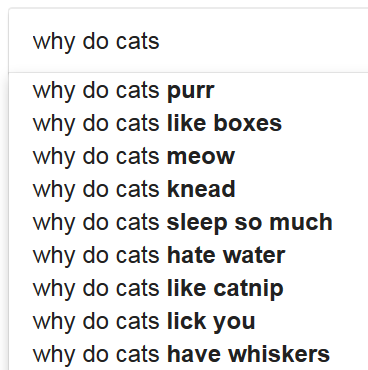}
\vspace*{-0.3cm}
    \caption{A glimpse into a search-engine query log.}
    \label{fig:querylogglimpse}
    \vspace*{-0.4cm}
\end{figure}

\vspace*{0.1cm}
\noindent{\bf Search engine logs}
Search engine logs 
%are arguably the most powerful 
are rich
collections of questions.
%For instance, projecting the number of questions in the AOL query dataset\footnote{https://en.wikipedia.org/wiki/AOL\_search\_data\_leak} that match the patterns we are interested in (\textasciitilde 0.09\%) into the traffic that Google receives daily (\textasciitilde 3.5 billion queries), the number of relevant questions asked to Google can be expected to be in the order of 3 million per day.
%
While logs 
%are well-guarded secrets, 
themselves are not available outside of industrial labs,
search engines allow us to glimpse at some of their
underlying statistics by auto-completion suggestions.
Figure~\ref{fig:querylogglimpse} shows an example of
this useful asset.
Quasimodo utilizes Google and Bing, which typically
return 5 to 10 suggestions for a given query prefix. 
%Both search engines normally return only 10/8 suggestions per input. 
In order to obtain more results, we recursively 
%and exhaustively 
probe the search engine with increasingly longer prefixes
that cover all letters of the alphabet, until 
the number of auto-completion suggestions drops
below 5. 
%less than 10/8 results are returned. For instance, if there are more than 10/8 suggestions 
For example, the query prefix \emph{``why do cats''}
is expanded into \emph{``why do cats a''}, \emph{``why do cats b''}, and so on. 

We intentionally restrict ourselves to query prefixes in
interrogative form, as these are best suited to convey
commonsense knowledge. In contrast, simple keyword queries
are often auto-completed with references to prominent entities
(celebrities, sports teams, product names, etc.), given the dominance
of such queries in the overall Internet
(e.g., the query prefix "cat" is expanded into "cat musical").
These very frequent queries are not useful for
CSK acquisition.

In total, we collected 11,603,121 questions from autocompletion.

\subsection{Question Patterns}
\label{subsec:patternextraction}

We performed a quantitative analysis of frequent question words and patterns on Reddit.
As a result, we decided to pursue two question words, \emph{Why} and \emph{How}, in combination with the verbs \emph{is, do, are, does, can, can't}, resulting in 12 patterns in total. Their relative frequency in the question set that we gathered by
auto-completion is shown in Table~\ref{tab:patterns_stats}. 
For forums, we performed title searches centered around these patterns.
For search engines, we 
appended subjects of interest to the patterns for
query generation (e.g., ``Why do cats'') for cats as subject.
The subjects were chosen from the common nouns extracted from WordNet \cite{Miller:1995:WLD:219717.219748}.

\begin{table}
    \centering
    \scalebox{0.85}{
    \begin{tabular}{ccc}
     \hline
     Pattern & In Query Logs & In QA Forums  \\
     \hline
     how does & 19.4\% & 7.5\% \\
     why is & 15.8\% & 10.4\%\\
     how do & 14.9\% & 38.07\%\\
     why do & 10.6\% & 9.21\%\\
     how is & 10.1 \% & 4.31\%\\
     why does & 8.97\% & 5.46\%\\
     why are & 8.68\% & 5.12\%\\
     how are & 5.51\% & 1.8\%\\
     how can & 3.53\% & 10.95\%\\
     why can't & 1.77\% & 1.40\%\\
     why can & 0.81\% & 0.36\%\\
     \hline
    \end{tabular}
    } % end scalebox
    \caption{Question patterns for candidate gathering.}
    \label{tab:patterns_stats}
    \vspace*{-0.1cm}
\vspace{-0.8cm}
\end{table}

%%%%%%%%%%%%%%%%%%%%%%%%%%%%%%%%%%%%%%%%%%%%%%%%%%%%%%%%%%%%%%

\subsection{From Questions to Assertions}
\label{subsec:openie}

%Given questions, our next challenge was to extract the actual beliefs or facts stated in them. As information extraction is typically aimed at assertional language, standard IE methods were not immediately applicable.
%We experimented with building custom extractors, but found this unexpectedly challenging, so in the end settled for transforming questions into statements, then applying standard OpenIE methods.

%one par on why this is different from standard IE
Open information extraction (Open IE) \cite{DBLP:conf/ijcai/Mausam16}, based on patterns,
has so far focused on assertive patterns applied to assertive sentences. 
%GW: assertive or asserting ?
In contrast, we deal with interrogative inputs,
facing new challenges.
%
%next par should state how we approach this
%Turning questions into assertions was in fact more difficult than expected as well. 

We address these issues by rule-based rewriting.
As we need to cope with colloquial or even ungrammatical language
as inputs, we do not rely on dependency parsing but merely
use part-of-speech tags for rewriting rules.
Primarily, rules remove the interrogative words and re-order
subject and verb to form an assertive sentence.
However, additional rules are needed to cast the sentence
into a naturally phrased statement that Open IE can deal with.
Most notably, auxiliary verbs like ``do'' need to be 
removed, and 
prepositions need to be put in their proper places,
as they may appear at different positions in interrogative
vs. assertive sentences.
Table~\ref{tab:question2statement} shows some example transformations, highlighting the modified parts. 

\begin{table}[t]
    \centering
    \scalebox{0.85}{
    \begin{tabular}{p{0.24cm} p{3.5cm} p{3.5cm}}
        \hline
         & Question & Statement \\
        \hline
        (1) & why \textit{is} voltmeter not connected in series & voltmeter \textit{is} not connected in series \\
        (2) & why \textit{are} chimpanzees endangered & chimpanzees \textit{are} endangered \\
        (3) & why \textit{do} men have nipples & men have nipples \\
        (4) & why \textit{are} elephant seals mammals & elephant seals \textit{are} mammals \\
        (5) & why \textit{is} becoming a nurse in france hard & becoming a nurse in france \textit{is} hard \\
        \hline
    \end{tabular}
    } % end scalebox
    \caption{Examples of questions and statements.}
    \label{tab:question2statement}
    \vspace{-0.9cm}
\end{table}

After transforming questions into statements, we employ the Stanford OpenIE
tool \cite{manning2014stanford} and OpenIE5.0 (\cite{saha-mausam-2018-open}, \cite{saha-etal-2017-bootstrapping},
\cite{pal2016demonyms}, \cite{christensen2011analysis})
 to extract triples from assertive sentences. 
When several triples with the same S and O are extracted from the same
sentence, we retain only the one with the longest P phrase.

%For very short and ungrammatical phrases,  OpenIE returns no results. 
%We overcome this limitation by applying a few hand-crafted extraction rules,
%turning phrases such as \emph{``lion roar''} and \emph{``lion can roar''} into the triple \fact{lion can roar}.

The resulting extractions are still noisy, but this is taken care of by the subsequent
stages of corroboration, ranking and grouping, to construct a high-quality CSKB.

%The output of this phase are quintuples \textit{(Subject, Predicate, Object, Module Source, Score)}, where the scores are source-specific, e.g. the position in the auto-complete suggestions, or OpenIE confidences. For instance, we have \textit{(lion, often hunts, zebra, Google, 0.4)}. 

%\sr{TODO: OpenIE confidence applies to all of them, so could be separated from source rank (Google/Bing), and is there a point to define a source rank for the forums as well?}

%\jr{For now, the score are directly combined, it makes things easier. For the forum, we could imagine the score are the upvotes (not done here)}

\subsection{Output Normalization}
\label{subsec:normalization}

The triples produced by OpenIE exhibit various idiosyncrasies. 
For cleaning them, we apply the following normalization steps:
%\begin{enumerate}
\squishlist
	\item Replacement of plural subjects with singular forms.
	\item Replacement of verb inflections by their infinitive
	(e.g., \fact{are eating} $\rightarrow$ \fact{eat}).
	\item Removal of modalities (always, sometimes, occasionally, ...) in the predicate and object. 
	These are kept as modality qualifiers.
	\item Removal of negation, put into a dedicated qualifier
	as negative evidence.
	\item Replacement of generic predicates like \fact{are}, \fact{is} with more specific ones like \fact{hasColor}, \fact{hasBodyPart}, depending on the %WordNet type of the
	object.
	\item Removal of adverbs and phatic expressions (e.g., ``so'', ``also'', ``very much'' etc.).
\squishend
%\end{enumerate}
%
We completely remove triples containing any of the following:
%\begin{enumerate}
\squishlist
	\item subjects outside the initial seed set,
	\item personal pronouns (``my'', ``we''), and
	\item a short list of odd objects (e.g., xbox, youtube, quote) that 
	frequently occur in search results but do not indicate commonsense properties.
\squishend
%\end{enumerate}

The output of this phase are tuples of the form \textit{(Subject, Predicate, Object, Modality, Negativity, Source, Score)}, for instance, \textit{(lion, hunts, zebra, often, positive, Google, 0.4)}.

%% file: parts/cleaningfusion.tex
\section{Corroboration}
\label{sec:cleaning}

%Large-scale question extraction is an important step towards achieving high coverage on human common-sense knowledge. Yet large-scale extraction inevitably returns also strange, obscure, and wrong assertions. The goal of the present step is to confirm the validity and relevance of the assertions extracted so far.
The output of the candidate gathering phase is bound to be
noisy and contains many false positives.
Therefore, we scrutinize the candidate triples by
obtaining corroboration signals from a variety of 
additional sources.
Quasimodo queries sources to test the occurrence and
obtain the frequency of SPO triples.
These statistics are fed into a logistic-regression
classifier that decides on whether a triple is accepted or not.
%Specifically, we build a supervised classification model for scoring candidate assertions, using the following additional web resources:

The goal of this stage is to validate whether candidate triples are plausible, i.e., asserting them is justified based on several corroboration inputs:
%\squishlist
%\item Is a triple meaningful and would be considered valid by most humans?
%\item Is a triple typical for its S arguments? That is, do most instantiations of S satisfy the PO property?
%\item Is a triple salient, in the sense that would most humans immediately associate the PO arguments with the
%given S argument?
%\squishend

%The goal of this step is to give a new score to the triples by using other data sources. Then, using a regression model trained with ConceptNet facts, we combine all the scores we have so far to validate the correctness of results from the previous step. Specifically, we train the model to predict whether a given fact is in ConceptNet, or not.

%However, ConceptNet only contains a dozen of relevant predicate so we can only evaluate the subject-object association. We can assume the predicates are correct for now as we assume the original questions were correct.

%We add the following features as input:
%For this corroboration, we tap into different kinds
%of sources so that all three dimensions are reflected.
%Specifically, we use the following sources:
%\begin{enumerate}
%\squishlist
%    \item English Wikipedia articles
%    \item Simple English Wikipedia
    %\item Antonym checking
%    \item Answer snippets from Google
%    \item Image tags from OpenImage\footnote{https://storage.googleapis.com/openimages}
%    \item Image tags from Flickr
%    \item Google Books
%    \squishend
%\end{enumerate}

\vspace{0.05cm}

\noindent{\bf Wikipedia and Simple Wikipedia.}
%These sources are used as follows: Given a fact candidate (S, P, O), 
For each SPO candidate,
we probe the article about S and compute the frequency of
co-occurring P and O within a window of $n$ successive words
(where $n=5$ in our experiments).
%compute the proportion of n-grams of the concatenation of P and O which occur in the Wikipedia page of S.
%
%\sr{More principled: One feature for P, one for O?}
%\jr{why not, but we have a lot of haveProperty. Still, it could be added}

%\paragraph{Antonym checking}
%Given a fact candidate (S, P, O), we check whether the set of fact candidates contains at least one candidate (S, P, O'), where WordNet lists O' as antonym of P. If that is the case, we set the corresponding feature to 1. E.g., if the candidates contain both (lion, is, dangerous) and (lion, is, peaceful)

\vspace{0.05cm}

\noindent{\bf Answer snippets from search engine.}
We generate Google queries using the S and O arguments of a triple as keywords, and analyze the top-100 answer snippets.
The frequency of snippets containing all of S, P and O
is viewed as an indicator of the triple's validity.
%count the frequency of predicates occurring in the top 50 answer snippets across four domains.
As search engines put tight constraints on the number of allowed
queries per day, we can obtain this signal only for a 
limited subset of candidate assertions. We prioritize 
the candidates for which the other sources (Wikipedia etc.)
yield high evidence.

\vspace{0.05cm}

\noindent{\bf Google Books.}
We create queries to the Google Books API
by first forming disjunctions of surface forms for each
of S, P and O, and then combining these into conjunctions.
For instance, for the candidate triple
(lion, live in, savanna), the query is "lion OR lions live OR lives in savanna OR savannas". As we can use the API only
with a limited budget of queries per day, we prioritized
candidate triples with high evidence from other sources
(Wikipedia etc.).

\vspace{0.05cm}

\noindent{\bf Image tags from OpenImages and Flickr.}
OpenImages is composed of ca. 20.000 classes used to annotate images. Human-verified tags exist for ca. 5.5 million images. 
Quasimodo checks for co-occurrences of S and O as
tags for the same image and computes the frequency of
such co-occurrences. 
%We only use the tag associations to confirm that S and O appear in same contexts. 
For Flickr, we use its API to obtain clusters of co-occurring tags.
%computed by Flickr, as the tags are not directly accessible. 
Individual tags are not available through the API.
We test for the joint occurrence of S and O in the same tag cluster.
%As for OpenImage, we give a score to a (S, P, O) triple based on the frequency of occurrence of S with O.

\noindent{\bf Captions from Google's Conceptual Captions dataset}
Google's Conceptual Captions dataset\footnote{\url{https://ai.google.com/research/ConceptualCaptions}} is composed of around 3 millions image descriptions. Using a method similar to the one used for Wikipedia, we check for a given fact SPO the concurrence of S with P and O.

%\sr{Other idea: For simple validation, check whether WP page of S contains O at least once, and vice versa? $\rightarrow$ could become another two binary features, and might allow to prune some odd facts, e.g., from jokes}

%\jr{It is hard to use jokes as most of the time, the fact is exacted but the answer is funny}
%\sr{My point was that this feature is specifically crafted to remove facts coming from jokes. E.g., there is a popular joke about elephants and cherry trees, but neither term occurs in the Wikipedia page of the other - maybe we could mention this here.}

\vspace*{0.15cm}
Table~\ref{tab:factprovenance} 
%shows how often each module was able to give a score.
gives the fractions of candidate triples for which each of the sources
contributes scoring signals.

\begin{table}
    \centering
    \scalebox{0.9}{
    \begin{tabular}{p{4cm}l}
        \hline
        Source & Fraction  \\
        \hline
        Google Auto-complete & 75.23\%\\
        Answers.com & 15.32\% \\
        Reddit & 11.16\% \\
        Yahoo! Answers & 1.17\% \\
        Quora, Bing Auto-complete & <1\% \\
        \hline
        CoreNLP Extraction & 63.21\%\\
        OpenIE5 Extraction & 51.24\%\\
        Custom Extraction & 10.79\%\\
        \hline
    \end{tabular}
    } % end scalebox
    \caption{Proportions of candidate triples by sources.}
    \label{tab:factprovenance}
    \vspace{-0.5cm}
\end{table}

%\begin{table}
%    \centering
%    \scalebox{0.9}{
%    \begin{tabular}{p{4cm}l}
%        \hline
%        Source & Fraction  \\
%        \hline
%        Wikipedia & 62.6\% \\
%        Simple Wikipedia & 53.6\% \\
%        Flickr & 2.72\% \\
%        OpenImage & 1.57\% \\
%        Quora, Bing Auto-complete, Google Books, Answer snippets & <1\% \\
%        \hline
%    \end{tabular}
%    } % end scalebox
%    \caption{Proportions of candidate triples by sources.}
%    \label{tab:factprovenance}
%    \vspace{-0.5cm}
%\end{table}

\vspace{0.05cm}

\noindent{\bf Classifier training and application.}
%
%We create positive training examples by assuming that fact candidates that are also in the human-created ConceptNet are correct (in total 1.75\% of our candidates). A similar number of negative samples is created by randomly sampling from the remaining candidates.
We manually annotated a sample of 700 candidate triples
obtained in the candidate gathering phase.\footnote{In comparison, TupleKB required crowd annotations for 70,000 triples.}
These are used to train a naive Bayes model, which gives us a precision of 61\%.

%% file: parts/ranking.tex
\section{Ranking}
\label{sec:ranking}

We refer to the scores resulting from the corroboration stage as plausibility scores $\pi$. 
These plausibility scores are essentially combinations of frequency signals. Frequency is an important criterion for ranking CSK, yet CSK has other important dimensions.

In this section we propose two probabilistic interpretations of the scores $\pi$, referred to as $\tau$ (``typicality'') and $\sigma$ (``saliency''). Intuitively, $\tau$ enables the ranking of triples by their informativeness for their subjects. Conversely, $\sigma$ enables the ranking of triples by the informativeness of their $p, o$ part.

To formalize this, we first define the probability of a triple $spo$. 
\[ \mathbf{P}[s , p, o] = \frac{\pi(spo)}{\Sigma_{x \in \textit{KB}}~~\pi(x)}. \]

Then, we compute $\tau$ and $\sigma$ as:

\begin{itemize}
    \item $\tau$(s, p, o) = $\mathbf{P}[p, o \mid s] = \frac{\mathbf{P}[s , p, o]}{\mathbf{P}[s]}.$
    \item $\sigma$(s, p, o) = $\mathbf{P}[s \mid p, o] = \frac{\mathbf{P}[s , p, o]}{\mathbf{P}[p, o]}.$
\end{itemize}

In each case, the marginals are $\mathbf{P}[p, o] = \Sigma_{s \in \textit{subjects}} ~\mathbf{P}[s, p, o]$ and $\mathbf{P}[s] = \Sigma_{p,o \in (\textit{predicates}, \textit{objects})} ~\mathbf{P}[s, p, o]$.

\vspace*{0.1cm}
At the end of this stage, each triple $spo$ is annotated with three scores: an internal plausibility score $\pi$, and two conditional probability scores $\tau$ and $\sigma$, which we subsequently use for ranking.

%(Do we use embeddings?)

%\sr{Unclear: On a per subject basis, $\pi$ and $\tau$ should give the same ranking. So how come that the evaluation results in Figure 3 show differences?} \sr{Analysis is rerun, possibly an artifact of missing row order randomization. But we may still consider this in the presentation, e.g., discard $\pi$ (intermediate score) in the diagrams, and only show $\tau$ and $\sigma$}

%% file: parts/refinementgrouping.tex
%\section{Refinement and Grouping}
\section{Grouping}
\label{sec:grouping}

%Given the use of text extraction in the pipeline so far, it is likely that resulting fact sets will contain some redundancy. In the present phase, we reduce the redundancy by grouping objects and predicates.

%\subsection{Object grouping}

The corroboration stage of Quasimodo aimed to remove
overly generic and overly specific assertions, but still
yields diverse statements of different granularities with
a fair amount of semantic redundancy.
For instance, \fact{hamsters are cute}, \fact{hamsters are cute pets}, and \fact{hamsters are cute pets for children} are all valid assertions,
but more or less reflect the same commonsense property.
Such variations occur with both O and P arguments, but less so with
the subjects S as these are pre-selected seeds in the candidate gathering
stage. 
%While the previous cleaning step aims to remove overly generic and overly specific facts, there is sometimes no clear best granularity. 
To capture such redundancies while preserving different granularities
and aspects, 
%yet not overload the KB with redundant information, 
Quasimodo groups assertions into near-equivalence classes.
At the top level, Quasimodo provides groups as entry points 
and then supports a meta-predicate \fact{refines} for
more detailed exploration and use-cases that need the full set
of diversely phrased assertions.
%we link such facts by a meta-predicate, \fact{refines}. 
%In the KB web interface, at the top level, only the most general facts per subject are displayed, along with information about their refinements.
%This \fact{refines} predicate is compute during the OpenIE phase, by looking at all the objects it yields for a given sentence. A question, in general, is transformed into several triples.

%\subsection{Co-clustering for predicate refinement}
%\subsection{Soft Co-Clustering}
%\label{subsec: coclustering}
\vspace*{0.1cm}
\noindent{\bf Soft Co-Clustering.}
%
%A major challenge for open information extraction are the variance among predicates, and overly generic predicates. For instance, 32\% of the TupleKB facts are made up by just 5 quite general predicates: \fact{have, isa, include, has-part} and \fact{is-part-of}. At the same time, 600 out of its 1600 predicates occur less than 10 times in the data. \jr{what about our generic predicates?} \kp{top-5 most freq. predicates in our data also cover approx. 40\% facts, but these predicates are also appeared in multiple clusters. Should we give more statistics on these top-5 or top-10 most freq. predicates?}
%
%Generic predicates and rare predicates are both challenges towards the usability of the data: Generic predicates may blur the specific semantic relationship between subject and object, while too rare predicates may inhibit the discovery of useful statistical patterns in the data. To specialize and normalize predicates, we propose to use biclustering of subject-object pairs and predicates. Such biclustering can achieve three goals: 
%
Our goal is to identify diverse formulations for both predicates P
and subject-object pairs SO.
The prior work on TupleKB has used ILP-based clustering to
canonicalize predicates. However, this enforces hard grouping
such that a phrase belongs to exactly one cluster.
With our rich data, predicates such as ``chase'' or ``attack''
can refer to very different meanings, though: predators chasing and attacking
their prey, or students chasing a deadline and attacking a problem.
Analogously, S and O arguments also have ambiguous surface forms
that would map to different word senses.

WebChild \cite{DBLP:conf/wsdm/TandonMSW14} 
has attempted to solve this issue by comprehensive
word sense disambiguation 
(see \cite{DBLP:journals/csur/Navigli09} for a survey), 
but this is an additional
complexity that eventually resulted in many errors. 
Therefore, we aim for the 
more relaxed and -- in our findings -- more appropriate
objective of computing {\em soft clusters} where the same phrase
can belong to different groups (to different degrees).
As the interpretation of P phrases depends on the context of
their S and O arguments, we cast this grouping task into
a {\em co-clustering} problem where SO pairs and P phrases are
jointly clustered.

\vspace*{0.1cm}
\noindent{\bf Tri-Factorization of SO-P Matrix.}
Our method for soft co-cluster\-ing of SO pairs and P phrases
is non-negative matrix tri-factorization;
see \cite{DBLP:conf/kdd/DingLPP06} for mathematical foundations.
We aim to compute clusters for SO pairs and clusters for P phrases
and align them with each other when meaningful.
For example, the SO pairs \fact{student\! problem} and
\fact{researcher\! problem} could be grouped together
and coupled with
a P cluster containing \fact{attack} and a second cluster containing
\fact{solve}.
This goal alone would suggest a standard form of factorizing
a matrix with SO pairs as rows and P phrases as columns.
However, the number of clusters for SO pairs and for P phrases
may be very different (because of different degrees of diversity in
real-world commonsense), and decomposing the matrix into two 
low-rank factors with the same dimensionality would not capture this
sufficiently well.
Hence our approach is tri-factorization where the number of (soft)
clusters for SO pairs and for P phrases can be different.

%to create $k$ and $l$ orthogonal SO and P soft clusters respectively. 
We 
%represent the retrieved SOP from the previous phase of our system pipeline in a matrix format, 
denote the set of SO pairs and P phrases, as observed in the
SPO triples after corroboration,
as an $m \times n$ matrix $M_{m\times n}$, where 
%an element $M_{ij}=score(spo)$ reflecting the saliency of the SPO triple, if the S-O pair from $i^{th}$ row co-occurs with P phrase from $j^{th}$ column. 
element $M_{ij}$ denotes the corroboration score of the triple with
$SO_i$ and $P_j$.
We factorize $M$ as follows:
$$M_{m\times n} = U_{m \times k} \times W_{k \times l} \times V^T_{~l \times n}$$
where the low-rank dimensionalities $k$ and $l$ are hyper-parameters
standing for the number of target SO clusters and target P clusters
and the middle matrix $W$ reflects the alignments between the two 
kinds of clusters.
The optimization objective in this tri-factorization 
is to minimize the data loss in terms of the Frobenius norm,
with non-negative $U,W,V$ and orthonormal $U,V$:

\vspace*{-0.2cm}
\begin{align}
   \textit{Minimize} \quad & \|M - U_{m \times k} \times W_{k \times l} \times V^T_{~l \times n} \|_{F} \nonumber \\ 
   s.t. \quad & U^TU = I, \quad V^TV = I \nonumber \\ 
   & U, V, W \geq 0 
   \label{eq_opt}
\end{align}
\vspace*{-0.2cm}

We can interpret $U_{i\mu}$ as a probability of 
the membership of the $i^{th}$ SO pair in the $\mu^{th}$ SO cluster. 
Similarly, $V_{j\nu}$ represents the probability of cluster membership of the $j^{th}$ P phrase to the $\nu^{th}$ P cluster. 
The coupling of SO clusters to P clusters is given by the $W_{k \times l}$ matrix, where the $\mu^{th}$ SO cluster is linked to the $\nu^{th}$ P cluster if $W_{\mu \nu} > 0$. 

\begin{table*}
    \centering
    \begin{tabular}{cccccccccc}\hline
        & &  &   & &  \multicolumn{2}{c}{\# SO/SO cluster}& \multicolumn{2}{c}{\# P/P cluster} & \# P clusters/P \\ 
    Domains & \#SPO & $k$ &  $l$ &$\rho$ &  avg. & max & avg. & max & avg.  \\ \hline
 
       Animals  & 201942 &  3500 & 2000 &0.10 & 38.46 & 383 & 2.8 & 24 & 1.5 \\
       Persons & 218924 & 5000 & 2000 & 0.10 & 11.7 & 235 &4.7 & 67 &  1.5 \\
       Medicine & 91184 & 3000 & 1800 & 0.15  & 45.17 & 171 & 2.91 & 31 & 1.3 \\
       Sport & 30794 & 1500 & 400 & 0.15 & 13.3 & 73  &3.8    & 15 & 1.14  \\ \hline
       macro-avg. (over & all 49 domains) & 1457.8 & 603.7 & 0.12 &33.97& 123.0& 3.5 & 24.8 & 1.24 \\ \hline
    \end{tabular}
    \caption{Statistics for 
    %empirically tuned clustering 
    SO clusters and P clusters for vertical domains Animals and Occupations.}
    \label{tab:hyper_param_clustering}
    \vspace{-0.5cm}
\end{table*}

\begin{table*}
    \centering
    \scalebox{0.9}{
    \begin{tabular}{p{6.5cm}|p{8cm}} \hline
         P clusters & SO clusters \\ \hline
         make noise at, be loud at, make noises at, croak in, croak at, quack at  & fox-night, frog-night, rat-night, mouse-night, swan-night, goose-night, chicken-night, sheep-night, donkey-night, duck-night, crow-night \\ \hline
        misbehave in, talk in, sleep in, be bored in, act out in, be prepared for, be quiet in, skip, speak in & student-class, student-classes, student-lectures   \\ \hline
        diagnose, check for & doctor-leukemia, doctor-reflexes, doctor-asthma, doctor-diabetes, doctor-pain, doctor-adhd \\ \hline
    \end{tabular}
    } % end scalebox
    \caption{Anecdotal examples of coupled SO clusters and P clusters from vertical domains Animals and Occupations.}
    \label{tab:example_clustering}
    \vspace{-0.6cm}
\end{table*}
%\kp{Should I explain more about these example clusters? }

%According to the problem setup, 
Each SO pair and P phrase have a certain probability 
of belonging to an SO and P cluster, respectively.
Hence, using a thresholding method, we assign ${SO_i}$ to 
the $\mu^{th}$ cluster if $U_{i\mu} > \theta$ and ${P_j}$ to the $\nu^{th}$ cluster if $V_{j\nu}>\theta$, in order to arrive at crisper clusters.
In our experiments, we set the thresholds as follows:
for the $\lambda^{th}$ SO cluster, we set $\theta_\lambda = \delta \cdot max_{i} U_{i\lambda}$, and for the $\lambda^{th}$ P cluster, we set $\theta_\lambda = \delta \cdot max_{i} V_{i\lambda}$. 
By varying the common thresholding parameter $\delta$, we tune the cluster assignments of SO pairs and P phrases based on the empirical perplexity
of the resulting clusters.
%We compare the perplexities of SO and P clusters, considering SO pairs and P phrases follow the multinomial probability distribution over their respective clusters. 
%Comparing the perplexity measure for $\delta = (0 , 0.5]$, we find the optimal $\delta = 0.1$ (better model holds lower perplexity). 
%This method of thresholding also produces less perplexity compare to fixed-rank thresholding for refining the cluster memberships.
This way, we found an empirically best value of $\delta = 0.1$.

The factor matrices in this decomposition should intuitively be sparse,
as each SO pair would be associated with only a few P clusters, and vice versa.
To reward sparsity, $L_1$ regularization
is usually considered for enhancing the objective function. 
However, the $L_1$ norm makes the objective non-differentiable,
and there is no analytic solution for the tri-factorization model.
Like most machine-learning problems, we rely on 
stochastic gradient descent (SGD) to 
approximately solve the optimization in Equation~\ref{eq_opt}. 
%Hence, we avoid $L_1$ regularization in our setup. 
For this reason, we do not use $L1$ regularization.
Our SGD-based solver initializes the factor matrices with a low density of
non-zero values, determined by a hyper-parameter $\rho$ for the ratio of
non-zero matrix elements.
The overall objective function then is the combination of
data loss and sparseness:
\vspace*{-0.05cm}
$$ \textit{Maximize}~~~ \frac{\textit{fraction~of~zero~elements}~(W)}{\textit{data~loss~by~Equation}~ \ref{eq_opt}}$$
\vspace*{-0.05cm}
All hyper-parameters -- the factor ranks $k$ and $l$ and the
sparseness ratio $\rho$ -- are tuned by performing a grid search.

%
%It indirectly forces many of the matrix elements to be assigned to near-zero values due to the characteristic of multiplicative update rules used in SGD, proposed by Ding et al.~\cite{DBLP:conf/kdd/DingLPP06}. 
%Such near-zero values are further removed by the thresholding method discussed earlier, resulting in sparse factored matrices.

%More specifically, our objective here is to find a solution of Equation~\ref{eq_opt} that ensures lower approximation error with higher sparsity of $W$ matrix (to computer crisper alignments between SO and P clusters). 
%Hence, in order to find the optimal hyper-parameters, $k, l,$ and $\rho$ for Equation~\ref{eq_opt}, we perform a grid search with objective function: $\max s(W)/\epsilon$, where $s(W)$ denotes sparseness of matrix $W$, i.e., $S(W) = 1- \frac{\text{\# nonzero-element}}{k\times l}$, and $\epsilon$ denotes factorization error. 

%% file: parts/experiments.tex
\section{Experimental Evaluation}

\subsection{Implementation}

\noindent{\bf Seeds.} As seeds for subjects we use a combination of concepts from ConceptNet, combined with nouns extracted from WordNet, 
% the most popular S concepts in ConceptNet: those that have at least 5 assertions, 
resulting in a total of around 120,000 subjects. 
%While our pipeline could also cope with more subjects, beyond these 17k, subjects are largely obscure and have little relation to common sense.
%Going beyond this size would entail huge noise with many exotic words.

\vspace{0.05cm}

\noindent{\bf Candidate Gathering.} In this phase %(Sec.~\ref{subsec:patternextraction}), from the QA forums, 
Quasimodo collected ca.\ 14,000 questions from Quora (which has
tight access restrictions), 
600,000 questions from Yahoo! Answers,
2.5 million questions from Answers.com (via its sitemap), and 3.5 million questions from a Reddit dump (with a choice of suitable sub-reddits). 
From auto-completion suggestions,
we obtained ca. 13 million questions from Google and 200,000 questions from Bing.
%After basic cleaning consisting in ignoring questions with personal words like ``I'' or ``your'' and questions which do not contain one of the required subjects, these numbers were consolidated to a total of around 5 million questions. 
%
%After transforming the questions into assertions, and applying OpenIE (Sec.~\ref{subsec:openie}), we retain 3.1 million fact candidates. After normalization (Sec.~\ref{subsec:normalization}), the number reduces to 2 million candidates. 
After applying the rewriting of questions into statements and running
Open IE, we obtained 7,5 million candidate triples; the subsequent
normalization further reduced this pool to ca.\ 2,3 million triples.

\vspace{0.05cm}

\noindent{\bf Corroboration.} The naive Bayes model assigned a mean $\pi$ score of 0.19, with a standard deviation of 0.22. For high recall we do not apply a threshold in this phase, but utilize the scores for ranking in our evaluations.
% Feature weights for the different sources are shown in Table  \ref{tab:coeffslogres}.

%{\tt\color{red} GW: we need to say something about this as well !!!!!} \sr{Put something, please check}

\vspace{0.05cm}

\noindent{\bf Grouping.}
We performed this step on the top-50\% triples, ordered by corroboration scores,
amounting to ca. 1 million assertions, 
%generated from the corroboration step. Before applying co-clustering method, 
For efficient computation, 
we sliced this data into 49 {\em basic domains} based on the WordNet domain hierarchy \cite{Bentivogli}. 
To this end, we mapped the noun sense of each assertion subject to WordNet and assign all triples for the subject to the respective domain (e.g., animals, plants, earth, etc.) 
The five largest domains are {\em earth, chemistry, animal, biology, and person}, containing on average 3.9k subjects and 198k assertions. 
We performed co-clustering on each of these slices, where hyper-parameters were tuned by grid search. 
%(Sec. \ref{sec:cleaning}). 
%Additionally, for in-depth evaluation,we computed clusters for two vertical domains, using top-50 most popular {\em animals} (e.g., cows, tigers, horses, etc.) and {\em occupations} (e.g., secretaries, judges, students, etc.). 
%
%%%%GW: the following is too much detail, compared to what we say about candidate gathering, hence commented out
%We leave out the SPO triples, where the P phrases or SO pairs co-occur with a single SO pair or P phrase, respectively. 
%For the co-clustering task on two vertical domains, we perform the grid search by varying the hyper-parameters, \# SO clusters ($k$) and P clusters ($l$), within $[50, min(\#SO, \#P)]$ with $step = 50$. The parameter $\rho$ is varied within $[0.1, 0.5]$. According to the grid search, the optimally tuned parameters for co-clustering, i.e., number of SO clusters and P clusters, are reported in Table~\ref{tab:hyper_param_clustering}. Additionally, we present different cluster-specific statistics, such as avg. and maximum size of SO and P clusters, avg. and max occurrences of a P phrase to different P clusters, and mean execution time of the co-clustering task in Table~\ref{tab:hyper_param_clustering}. After co-clustering the filtered triples, the space of P phrases is reduced by 67\% and 63\% respectively for the verticals on animals and occupations. Samples from resulting coupled SO- and P-clusters are presented in Table~\ref{tab:example_clustering}. 
%
Table~\ref{tab:hyper_param_clustering} gives hyper parameter values and cluster-specific statistics of the co-clustering for three domains:
number of assertions (\#SPO); the co-clustering hyper-parameters
SO clusters (k), P clusters (l) and sparseness ratio ($\rho$); 
the average number of elements per cluster for both SO and P clusters; and 
the average number of P-clusters per predicate. 
Additionally, we provide macro-averaged statistics for all 49 domains. 
Table~\ref{tab:example_clustering} shows anecdotal examples of co-clusters for illustration.

\vspace{0.05cm}

\noindent{\bf Quasimodo CSKB.}
The resulting knowledge base contains ca.\
2.3 million assertions for 80,000 subjects.
Quasimodo is accessible online.\footnote{\url{https://www.mpi-inf.mpg.de/departments/databases-and-information-systems/research/yago-naga/commonsense/quasimodo/}}
%The total size of our resulting KB is shown in Table~\ref{tbl:size}, where we also report statistics for related projects. As one can see, Quasimodo contains significantly more statements per subject than all KBs but WebChild.

\vspace{0.05cm}

\noindent{\bf Run-Time.} One of the expensive component of
Quasimodo is the probing of Google auto-completions. 
This was carried out within the allowed query limits
over an entire week. Bing auto-completion was accessed through the Azure API. %For forums, a dump of Reddit was used, while Quora was accessed semi-manual, and questions from Answers.com were extracted from the sitemap.
Another expensive component is co-clustering of all 49 domains, which takes total 142 hours in a  Intel Xeon(R)(2 cores@3.20GHz) server (average 3.14 hours/ slice).
All other components of Quasimodo run within a few hours at most.

%\begin{itemize}
%    \item 18k subjects as seeds
%    \item 13k subjects with facts
%    \item 1.3 million facts
%    \item 344k objects after doing some basic normalization
%    \item 89k predicates
%    \item 5.8k predicates with more than 10 facts cover 1.2 million of facts
%\end{itemize}

\begin{table*}
\centering
\scalebox{0.9}{
\begin{tabular}{@{}crrrrr@{}}
\toprule
 \multicolumn{6}{c}{Full KB} \\
                & \#S & \#P & \#P$\geq$10 & \#SPO & \#SPO/S \\ \midrule
ConceptNet-full$@$en &  842,532  & 39   & 39  & 1,334,425 & 1.6       \\ 
ConceptNet-CSK@en  &   41,331   & 19   &  19    &  214,606  & 5.2    \\ 
TupleKB         & 28,078   & 1,605   & 1,009  & 282,594 & 10.1   \\    %Lock this, verified, TupleKB's subjects are not normalized but we'll just use it as it is
WebChild        &  55,036   & 20  & 20   &  13,323,132  & 242.1   \\
\textbf{Quasimodo} & 80,145 & 78,636 & 6084 & 2,262,109 & 28.2   \\ \bottomrule
\end{tabular}
\qquad
\begin{tabular}{@{}crrrr@{}}
\toprule
\multicolumn{2}{c}{animals} & \multicolumn{2}{c}{occupations} \\ 
                \#S & \#SPO & \#S &   \#SPO \\ \midrule
%ConceptNetfull 50 &     5,756     & 50  &    2,365         \\
 50 &     2,678     & 50  &    1,906         \\ %Lock this, verified
 50     &  1,841         & 50    &  1,495       \\
%TupleKB 49  &  16,052  & 38 &  5,321  \\
 49  &  16,052  & 38 &  5,321  \\              %Lock this, verified, 

 50  &  27,223  & 50 & 26,257    \\
50  & 39,710   & 50 & 18,212    \\ \bottomrule
\end{tabular}
}
\caption{Statistics for different KBs. Left side full KBs, right side two slices on animals and occupations.}
\label{tbl:size}
\vspace{-0.6cm}
\end{table*}

\subsection{Intrinsic Evaluation}

We evaluate four aspects of Quasimodo: 1) size of the resulting CSKB, 2) quality, 3) recall, and 4) cluster coherence.

%\subsubsection{Size}

\vspace*{0.1cm}
\noindent{\bf Size.}
We compare KBs in Table~\ref{tbl:size} left side 
by the number of subjects (\#S), the number of predicates (\#P),
predicates occurring at least 10 times (\#P$\geq$10), and the number of triples (\#SPO).
For Quasimodo we exclude all triples with \fact{isA / type} predicate
denoting subclass-of or instance-of relations, as these are well covered in traditional knowledge resources like WordNet, Wikidata and Yago.
We also compare, on the right side of the table,
on two vertical domains: assertions for
the 50 most popular animals and 50 most popular occupations, as determined by frequencies from Wiktionary. 
%(Table~\ref{tbl:size} right).
For ConceptNet, we report numbers for the full data including 
 \fact{isA / type} and 
\fact{related} and other linguistic triples (e.g., on etymology) 
imported from DBpedia, WordNet and Wiktionary (ConceptNet-full), and for the proper CSK core where these relations are removed (ConceptNet-CSK).

Table~\ref{tbl:size} clearly conveys that
Quasimodo has richer knowledge per subject than all other resources
except for WebChild. The advantage over the manually created ConceptNet becomes particularly evident when looking at the two vertical domains, where ConceptNet-CSK contains less than 10\% of the assertions that Quasimodo knows.

\vspace*{0.1cm}
%\subsubsection{Precision}
\noindent{\bf Quality.}
We asked MTurk crowd workers to evaluate the 
%precision 
quality
of CSK assertions
along three dimensions: 1) meaningfulness, 2) typicality, 3) saliency.
Meaningfulness denotes if a triple is conveys meaning at all, or is absurd;
typicality denotes if most instances of the S concept have the PO property;
saliency captures if humans would spontaneously associate PO with the
given S as one of the most important traits of S.

For each evaluated triple, 
we obtained two judgments for the three aspects, each graded on a scale from 1 (lowest) to 5 (highest). 
A total 275 crowd workers completed the evaluation, with mean variance 0.70 on their ratings 
from 1 to 5 indicating good inter-annotator agreement.
%The aspects were whether 1) the facts were comprehensible, 2) whether they would apply to most S-instances, and 3) whether the fact was relevant or interesting. 
%To ensure that crowd workers would not be distracted by KB-specific jargon, we translated predicates like \fact{hasPrerequisite} or \fact{hasProperty} into generic verb phrases like ``requires'' or ``is''. 

We sampled triples from the different CSKBs under two settings:
In \emph{comparative sampling}, we sampled triples for the same 100 subjects (50 popular occupations and 50 popular animals) across all KBs. 
For subject and each KB we considered the top-5-ranked triples as a pool, and uniformly sampled
100 assertions for which we obtain crowd judgement.
For Quasimodo, as the rankings by typicality $\tau$ and by saliency $\sigma$ differ, this sampling
treated Quasimodo-$\tau$ and Quasimodo-$\sigma$ as distinct CSKBs.
This setting provides side-by-side comparison of triples for the same subjects.

In \emph{horizontal sampling}, we sampled each KB separately; so they could differ on the evaluated
subjects. We considered the top 5 triples of all subjects present in each KB as a pool,
and picked samples from each KB uniformly at random.
This evaluation mode gave us insights into the average quality of each KB.
Note that it gives KBs that have fewer long-tail subjects an advantage, as triples for long-tail subjects 
usually receive lower human scores. 
Again, we considered Quasimodo rankings by $\tau$ and $\sigma$ as distinct CSKBs.
%thus compare a total of 5 KBs.

\begin{figure}
\begin{center}
\begin{tikzpicture}
\begin{axis}[
    ybar,
    ymin=2.5, ymax=4.5,    
    %cycle list name=color list,
    bar width=0.2cm,
    height=3cm,
    width=\columnwidth,
    enlarge x limits=0.3,
    ylabel={quality@5},
    symbolic x coords={meaningfulness,typicality,saliency},
    ylabel near ticks, 
    xtick=data,
    legend style={at={(0.5,-0.6)},
    anchor=north,legend columns=-1},
    ]
\addplot[black,fill=yellow,pattern=north east lines] coordinates {(meaningfulness, 3.98) (typicality, 3.7) (saliency,3.52)};
\addplot[olive,fill=olive] coordinates {(meaningfulness,3.23) (typicality,3.02) (saliency,2.8)};
\addplot[teal,fill=teal] coordinates {(meaningfulness, 3.77) (typicality,3.64) (saliency,3.42)};
%\addplot[blue,fill=blue] coordinates {(meaningfulness,3.34) (typicality, 3.20) (saliency,3.14)}; % remove this again!
\addplot[black,fill=blue!25!white] coordinates {(meaningfulness,3.84) (typicality,3.69) (saliency, 3.1)};
\addplot[black,fill=blue!75!black] coordinates {(meaningfulness, 2.93) (typicality, 2.63) (saliency,3.3)};

\legend{ConceptNet, WebChild ,TupleKB, Q'modo-$\tau$, Q'modo-$\sigma$};
\end{axis}
\end{tikzpicture}
\vspace*{-0.7cm}
\caption{Quality for horizontal sampling.}
\label{fig:eval-correctness-new-2}
\end{center}
\vspace{-0.4cm}
\end{figure}
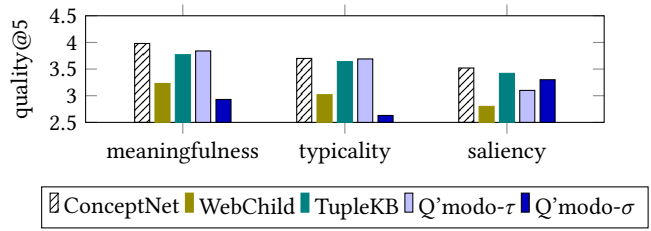

The results of these evaluations are shown in Figure~\ref{fig:eval-correctness-new-1} and Figure~\ref{fig:eval-correctness-new-2}.  
With comparative sampling, Quasimodo-$\tau$ significantly outperforms both WebChild and TupleKB, 
%(+0.9 -- +1.1), 
and nearly reaches the quality of the human-generated ConceptNet. 
In horizontal sampling mode, Quasimodo-$\tau$ outperforms WebChild along all dimensions 
%(+0.3 -- +0.6), 
and outperforms TupleKB in all dimensions but saliency.
This is remarkable given that Quasimodo is 3 times bigger than ConceptNet, 
and is therefore penalized with horizontal sampling by its much larger number of long-tail subjects.
In both evaluations, Quasimodo-$\tau$ significantly outperforms Quasimodo-$\sigma$ in terms of meaningfulness and typicality. Regarding saliency the results are mixed, suggesting that further research on ranking models would be beneficial.

\begin{figure}
\begin{center}
\begin{tikzpicture}
\begin{axis}[
    ybar,
    ymin=2.5, ymax=4.5,    
    %cycle list name=color list,
    bar width=0.2cm,
    height=3cm,
    width=\columnwidth,
    enlarge x limits=0.3,
    ylabel={quality@5},
    symbolic x coords={meaningfulness,typicality,saliency},
    ylabel near ticks, 
    xtick=data,
    legend style={at={(0.5,-0.6)},
    anchor=north,legend columns=-1},
    ]
\addplot[black,fill=yellow,pattern=north east lines] coordinates {(meaningfulness, 4.11) (typicality, 3.96) (saliency,3.77)};
\addplot[olive,fill=olive] coordinates {(meaningfulness, 2.99) (typicality,2.66) (saliency,2.53)};
\addplot[teal,fill=teal] coordinates {(meaningfulness, 2.93) (typicality, 2.83) (saliency, 2.54)};
%\addplot[blue,fill=blue] coordinates {(meaningfulness,3.84) (typicality, 3.62) (saliency,3.33)};
\addplot[black,fill=blue!25!white] coordinates {(meaningfulness,3.98) (typicality,3.70) (saliency, 3.43)};
\addplot[black,fill=blue!75!black] coordinates {(meaningfulness, 3.5) (typicality, 3.12) (saliency,3.08)};

\legend{ConceptNet, WebChild ,TupleKB, Q'modo-$\tau$, Q'modo-$\sigma$};
\end{axis}
\end{tikzpicture}
\vspace{-0.6cm}
\caption{Quality for comparative sampling.}
\label{fig:eval-correctness-new-1}
\end{center}
\vspace{-0.6cm}
\end{figure}
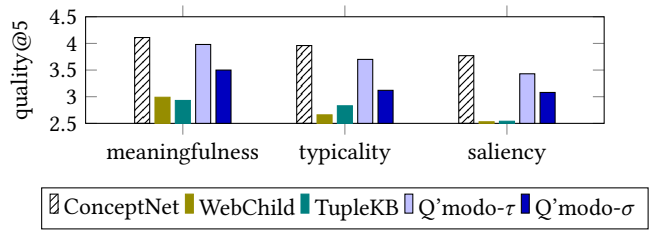

\begin{figure}
% part 1: full only
    \includegraphics[width=\columnwidth]{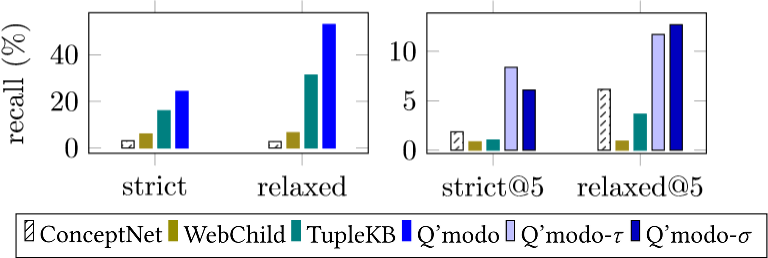}
\vspace{-0.8cm}
\caption{Recall evaluation.}
\label{fig:eval-recall}
\vspace{-0.6cm}
\end{figure}

\begin{table*}
\centering
\scalebox{0.8}{
\begin{tabular}{@{}llll@{}}
\toprule
%\multicolumn{4}{c}{Elephant} \\ \midrule
\multicolumn{1}{c}{Quasimodo} & \multicolumn{1}{c}{ConceptNet} & \multicolumn{1}{c}{WebChild} & \multicolumn{1}{c}{TupleKB} \\ \midrule
(hasPhysicalPart, trunk) & (AtLocation, africa) & (quality, rare) & (has-part, brain) \\
(hasPhysicalPart, ear) & (HasProperty, cute) & (trait, playful) & (drink, water) \\
(live in, zoo) & (CapableOf, remember water source) & (size, large) & (prefer, vegetation) \\
(love, water) & (HasProperty, very big) & (state, numerous) & (eat, apple) \\
(be in, circus) & (CapableOf, lift logs from ground) & (quality, available) & (open, mouth) \\ \bottomrule
\end{tabular}
}

\vspace{0.15cm}

\scalebox{0.8}{
\begin{tabular}{@{}llll@{}}
\toprule
%\multicolumn{4}{c}{Doctor} \\ \midrule
\multicolumn{1}{c}{Quasimodo} & \multicolumn{1}{c}{ConceptNet} & \multicolumn{1}{c}{WebChild} & \multicolumn{1}{c}{TupleKB} \\ \midrule
(help, people) & (HasA, private life) & (emotion, euphoric) & (complete, procedure) \\
(stand long for, surgery) & (CapableOf, attempt to cure patients) & (quality, good) & (conduct, examination) \\
(learn about, medicine) & (AtLocation, golf course) & (trait, private) & (get, results) \\
(cure, people) & (CapableOf, subject patient to long waits) & (atlocation, hospital) & (has-part, adult body) \\
(can reanimate, people) & (AtLocation, examination room) & (hasproperty, aggressive) & (treat, problem) \\ \bottomrule
\end{tabular}
}
\caption{Anecdotal examples (PO) for S \fact{elephant} (top) and S \fact{doctor} (bottom).}
\label{tbl:casestudy}
\vspace{-0.6cm}
\end{table*}

\vspace*{0.1cm}
\noindent{\bf Recall.}
To compare the recall (coverage) of the different CSKBs, 
we asked crowd workers at MTurk to make statements about 50 occupations and 50 animals as subjects. We asked to provide short but general sentences, as spontaneous as possible so as to focus on typical and salient properties.
Together with these instructions, 
we gave three examples for elephants (e.g., ``elephants are grey'', ``elephants live in Africa'') and three examples for nurses. 
For each subject, crowd workers had 4 text fields to complete, which were pre-filled with ``[subject] ...''. Each task was handed out 6 times;
so in total, we obtained 2,400 simple sentences on 100 subjects.

\input{parts/qa_results_ml.tex}

We computed CSKB recall w.r.t.\ these crowd statements in two modes.
In the {\em strict} mode,  we checked for each sentence if the KB contains a triple (for the same subject) where both predicate and object are contained in the sentence and, if so, computed the word-level token overlap 
between PO and the sentence.
%weighing the match by the overall token overlap. 
In the {\em relaxed} setting, we checked separately if the KB contains an S-triple whose predicate appears in the sentence, and if it contains an S-triple whose object appears in the sentence. The results are shown in  Figure~\ref{fig:eval-recall}. 
In terms of this coverage measure, 
Quasimodo  outperforms the other CSKBs by a large margin,
in both strict and relaxed modes
%both when using the full KB and when using only top 5 facts per subject, with the larger difference in the top-5 setting indicating the quality of its scoring function.
and also when limiting ourselves to the top-5 highest-ranked
triples per subject.

\vspace{-0.1cm}

\vspace*{0.1cm}
%\subsubsection{Cluster coherence}
\noindent{\bf Cluster Coherence.}
We evaluate cluster coherence using an intruder task. For a random set of clusters that contain at least three P phrases, we show 
annotators 
%3-5
%up to 5 
sampled SO pairs from the cluster and 
%3 
samples of P phrases 
from the aligned cluster interspersed 
with an additional random intruder predicate 
drawn from the entire CSKB. 
For example, we show the SO pairs \fact{spider-web, mole-tunnel, rabbit-hole}, along with the P phrases \fact{build, sing, inhabit, live in}, where \fact{sing} is the intruder to be found. We sampled 175 instances from two vertical slices, Animals and Persons, and used crowdsourcing (MTurk) to collect a total of 525 judgments on these 175 instances for the intruder detection task. We obtained an intruder detection accuracy of 64\% for clusters in the Animals domain, and 54\% in Persons domain (compared with 25\% for a random baseline). 
This is supporting evidence that our co-clustering method yields
fairly coherent groups.
%This likely reflects the more diverse predicates existing for animals.
%%%GW: commented the last sentence out -- very hard to interpret this situation

%\vspace{-0.15cm}

%\subsubsection{Anecdotal examples}
\vspace*{0.1cm}
\noindent{\bf Anecdotal Examples.}
Table~\ref{tbl:casestudy} provides a comparison of randomly
chosen assertions for two subjects in each of the KBs:
\fact(elephant) (top) and \fact{doctor} (bottom).
WebChild assertions are quite vague, while TupleKB assertions are reasonable but not always salient. 
ConceptNet, constructed by human crowdsourcing, features high-quality  assertions, but sometimes gives rather exotic properties.
In contrast, the samples for Quasimodo are
both typical and salient.

%\balance

\vspace{-0.2cm}
\subsection{Extrinsic Evaluation}

\input{parts/qa_explanation_ml.tex}

\noindent{\bf Word Guessing Game.}
Taboo is a popular word guessing game in which a player describes a 
concept without using 5 taboo words, usually the strongest cues. 
The other player needs to guess the concept.
We used a set of 578 taboo cards from the website \textit{playtaboo.com} to evaluate the coverage of the different CSKBs.

%\begin{table}
%\begin{tabular}{ccc}
%\hline
%KB & \begin{tabular}{c}Taboo words\\ (via P or O)\end{tabular} & \begin{tabular}{c}Taboo words\\ (via O only)\end{tabular} \\ \hline
%Quasimodo & \textbf{31.4\%} & \textbf{29.3\%}\\
%ConceptNet & 26.1\% & 26.1\%\\
%WebChild & 20.0\% & 19.7\%\\
%TupleKB & 11.3\% & 10.0\%\\ \hline
%Quasimodo1.1 & \textbf{34.2\%} & \textbf{32.1\%}\\
%\end{tabular}
%\caption{Coverage for word guessing game.\sr{Plot as graph}}
%\vspace{-0.9cm}
%\end{table}

\begin{figure}
\begin{center}
\begin{tikzpicture}
\begin{axis}[
    ybar,
    bar width=0.2cm,
    height=3cm,
    width=\columnwidth,
    enlarge x limits=0.65,
    legend style={at={(0.5,-0.35)},
    anchor=north,legend columns=-1},
    ylabel={Coverage (\%)},
    ylabel near ticks,
    symbolic x coords={via P or O,via O only},
    xtick=data,
    ymin=0,
    ]
\addplot[black,fill=yellow,pattern=north east lines] coordinates {(via P or O,39.7) (via O only, 39.6) };
\addplot[olive,fill=olive] coordinates {(via P or O,33.4) (via O only,32.8) };
\addplot[teal,fill=teal] coordinates {(via P or O,18.2) (via O only,16.8) };
\addplot[blue,fill=blue] coordinates {(via P or O,45.3) (via O only,44.4) };
\legend{ConceptNet,WebChild,TupleKB,Quasimodo}
\end{axis}
\end{tikzpicture}
\vspace{-0.3cm}
\caption{Coverage for word guessing game.}
\label{tbl:taboo}
\end{center}
\vspace{-0.5cm}
\end{figure}
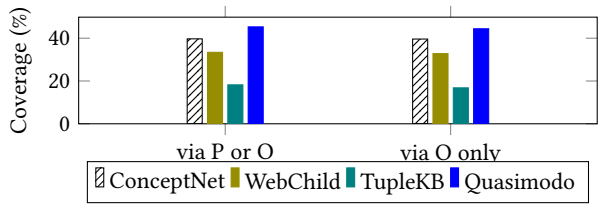

Given a concept to be guessed, 
we compute the fraction of Taboo words that a KB associates with
the concept, 
appearing in the O or P argument of the triples for the concept.
This is a measure of a CSKB's potential ability to perform
in this game (i.e., not playing the game itself).
The resulting coverage is shown in Table~\ref{tbl:taboo}. 
Quasimodo outperforms all other KBs by this measure.
TupleKB, the closest competitor on the science questions in the multiple-choice QA use case, 
has substantially lower coverage, 
indicating its limited knowledge beyond the (school-level) science domain.

%\subsubsection{Association prediction}

%We use questions and answers from the game Family Feud to evaluate how well our KB can discover word associations. Family Feud is a game where teams predict associative knowledge, for instance, ``Name something you might be asked to bring to a friend's party.''. Teams then have to predict the top 3-6 answers that 100 random people gave (here e.g., food, drinks, guests, chairs, music, ..). We used answers to this game to evaluate how well we could predict terms in the questions, i.e., given \emph{food, drinks, guests, chairs, music}, tried to predict words in the question, like \emph{party}

%\sr{Check whether that is done only for informative terms}

%The results are as follows: 

%TODO

%% file: parts/qa_results_ml.tex
\begin{table*}
    \centering
    \scalebox{0.8}{
    \begin{tabular}{ccccccc}
        \hline
        KB & Elementary NDMC & Middle NDMC & CommonsenseQA2 & Trivia & Examveda & All \\
        \hline
        \#Questions (Train/Test) & 623/541 & 604/679 & 9741/1221 & 1228/452 & 1228/765 & 10974/3659\\ \hline
        Random & 25.5 & 23.7 & 21.0 & 25.9 & 25.4 & 22.0 \\
        word2vec & 26.2 & 28.3 & 27.8 & 27.4 & 25.6 & 27.2 \\
        %ElasticSearch & 45/45 & 41.8/41.8 & 38.2/38.2 & 31.6/31.6 & 37.9/37.9 \\ 
\hdashline
        Quasimodo & \textbf{38.4} & \textbf{34.8} & 26.1 & \textbf{28.1} & \textbf{32.6} & \textbf{31.3} \\
        ConceptNet & 28.5 & 26.4 & \textbf{29.9} (source) & 24.4 & 27.3 & 27.5\\
        TupleKB & 34.8 & 25.5 & 25.3 & 22.2 & 27.4 & 27.5 \\
        WebChild & 26.2 & 25.1 & 25.2 & 25.9 & 27.1 & 24.1 \\
    \hline
    \end{tabular}
    }
    \caption{Accuracy of answer selection in question answering.}
    %Statistically significant results (with p-value $< 0.05$ for paired t-test) of Quasimodo against other CSKBs are marked with an asterisk.}
    \label{tbl:QA}
    \vspace{-0.6cm}
\end{table*}
% 38.0/33.3 scores of conceptnet on commonsenseQA

%% file: parts/qa_explanation_ml.tex
\noindent{\bf Answer Selection for QA.}
In this use case, we show that CSK helps in selecting answers
for multiple-choice questions. We use five datasets: (i+ii) elementary school and middle school science questions from the AllenAI science challenge~\cite{ai2-dataset}, (iii) commonsense questions generated from ConceptNet~\cite{commonsenseqa}, (iv) reading comprehension questions from the TriviaQA dataset~\cite{triviaqa}, and (v) exam questions from the Indian exam training platform Examveda~\cite{parchureveda}. The baseline are given by AllenAI, in aristo-mini\footnote{\url{https://github.com/allenai/aristo-mini}}.

To assess the contribution of CSKBs, for each question-answer pair
we build a pair of contexts $(ctx_{\textit{question}}, ctx_{\textit{answer}})$ as follow:
for each group of words \textit{grp} in the question (resp. the answer),
we add all the triples of the form \textit{(grp, p, o)} or \textit{(s, p, grp)} in a KB.
Then, we create features based on $(ctx_{\textit{question}}, ctx_{\textit{answer}})$ such as the number of SPO, SP, SO, PO, S, P, O overlaps, the size of the intersection of a context with the opposition original sentence, the number of useless words in the original sentence and the number of words in the original sentences. From these features, we train an AdaBoost classifier.
This is a basic strategy for multiple-choice QA and could
 be improved in many ways. However, it is sufficient to
bring out the value of CSK and the differences between the CSKBs.

We compare four CSKBs against each other and
against a word2vec baseline which computes the embeddings
similarity between questions and answers.
%in order to assess where the common-sense facts add more than just association knowledge.
The results are shown in Table~\ref{tbl:QA}. 
%As one can see, 
Quasimodo significantly outperforms the other CSKBs on four of the five datasets (ConceptNet performs better on CommonsenseQA dataset as it was used for the generation).

%% file: parts/conclusion.tex
\vspace*{-0.1cm}
\section{Conclusion}

This paper presented Quasimodo, a methodology for acquiring
high-quality commonsense assertions, 
by harnessing non-standard input sources, like
query logs and QA forums, in a novel way.
%relying on powerful web resources, in particular, query logs and answer snippets. 
%
%As Quasimodo can be flexibly adapted to different inputs, it is well suited to generate common sense assertions on a range of domains and provides a powerful resource for tasks such as question answering and dialogue.
%
%%%GW:re-emphasize our USPs - unique sales points
As our experiments demonstrate, the Quasimodo knowledge base
improves the prior state of the art, by achieving much better
coverage of typical and salient commonsense properties (as determined by
an MTurk study) while having similar quality in terms of precision.
Extrinsic use cases further illustrate the advantages of Quasimodo.
%In Quasimodo, we have used general English-language web resources. An interesting extension would be to extract data from country-specific websites and domains, thus identifying regional and cultural common-sense and preconceptions.
%
The Quasimodo data is available online\footnote{\url{https://www.mpi-inf.mpg.de/departments/databases-and-information-systems/research/yago-naga/commonsense/quasimodo/}}, and our code is available on Github\footnote{\url{https://github.com/Aunsiels/CSK}}.

\vspace{0.1cm}
\noindent
\textbf{Acknowledgment.}
We thank the anonymous reviewers and Niket Tandon for helpful comments.

%% file: parts/appendix.tex
\section{Quasimodo v4.3 (2/2021)}

\subsection{Subject Augmentation}

After the normalisation phase, we only retain subjects that are in a defined list. This filtering allows us to discard esoteric subjects, as potential subjects are quasi-infinite and often, do not generalise to a more global context. For example, we removed \textit{dancing nun}, \textit{video help} and \textit{Twitter user type}. Note that the original filter also removed subjects that could be interesting to consider, but are not present in ConceptNet or WordNet. We give here some examples of potentially useful subjects:
\begin{itemize}
    \item Fictional characters such as \textit{John Snow} or \textit{Harry Potter}
    \item Notorious people such as \textit{Donald Trump} or \textit{Hillary Clinton}
    \item Actions such as \textit{eating asparagus} (thus, we can extend ConceptNet actions) or \textit{drinking tea}
    \item Objects such as \textit{energy drink} or \textit{mechanical keyboards}
    \item Concepts such as \textit{gas prices} or \textit{social media}
    \item etc.
\end{itemize}

These subjects are of primary interest as people often talk about them. So, we extract them using the following procedure:
\begin{itemize}
    \item During the subject removal step, count for each ignored subject how often it appears.
    \item Filter ignored subjects that appear less than a certain threshold (10 appearances in our experiments)
    \item Filter subjects that contain personal words (such as \textit{my}, \textit{your} and \textit{its})
    \item Filter subjects that start by a determinant (such as \textit{an elephant ear})
\end{itemize}

Finally, we get a list of new 148k subjects that the pipeline is going to use during the next iteration.

\subsection{Additional Patterns}

We kept increasing the number of question patterns allowed by the system. In Table~\ref{tab:additional_patterns_stats}, we present the current patterns and their frequency in the final generated knowledge base. It is important to see that some of these patterns allow extracting negative knowledge.

\begin{table}
    \centering
    \begin{tabular}{cc}
     \hline
     Pattern & Frequency\\
     \hline
        why is <SUBJ> & 30\%\\
        why are <SUBJS> & 21\%\\
        why do <SUBJS> & 10\%\\
        how is <SUBJ> & 8\%\\
        how is a <SUBJ> & 5\%\\
        why does <SUBJ> & 5\%\\
        how are <SUBJS> & 5\%\\
        how does <SUBJ> & 4\%\\
        how do <SUBJS> & 3\%\\
        why is a <SUBJ> & 2\%\\
        why isn't <SUBJ> & 2\%\\
        how can <SUBJS> & 1\%\\
        why aren't <SUBJS> & 1\%\\
        why <SUBJ> & 1\%\\
        why don't <SUBJS> & <1\%\\
        why can <SUBJS> & <1\%\\
        why can't <SUBJS> & <1\%\\
        how does a <SUBJ> & <1\%\\
        how can a <SUBJ> & <1\%\\
        why doesn't <SUBJ> & <1\%\\
        why isn't a <SUBJ> & <1\%\\
        why does a <SUBJ> & <1\%\\
        why can't a <SUBJ> & <1\%\\
        why can a <SUBJ> & <1\%\\
        why doesn't a <SUBJ> & <1\%\\
     \hline
    \end{tabular}
    \caption{Question patterns for candidate gathering}
    \label{tab:additional_patterns_stats}
\end{table}

\subsection{Data Slices}

As Quasimodo is getting bigger and is useful for several communities with different goals, we decided to generate slices of Quasimodo on specific themes. We give in Table~\ref{tab:domain_samples} statistics about these samples. First, we separated positive and negative facts. Then, we isolated the top 10\% facts to produce a dataset with better overall precision. Next, we created a dataset for cultural statements. More precisely, we only kept statements where a country or its inhabitants (extracted from a list of 112 countries) appear in the subject or the object. The same way, we created a dataset of animals and occupations.

As ConceptNet is a popular resource for commonsense knowledge and many applications use its scheme (composed of very few relations), we reduced Quasimodo to ConceptNet relations. This operation can be very complicated. Therefore, we did it by manually writing simple rules that allow capturing simple conversions. Some relations barely appear in Quasimodo (particularly the ones about events, which are also sparse in ConceptNet), so we ignored them.

\begin{table}[]
    \centering
    \begin{tabular}{|l|c|c|c|}
        \hline
        Quasimodo Version & \#S & \#P & \#SPO \\
        \hline
        Quasimodo v1.0 & 80,145 & 78,636 & 2,262,109 \\
        \hline
        Quasimodo v4.3 & 148,627 & 128,638 & 6,281,580 \\
        Positive & 145,020 & 125,241 & 5,930,628 \\
        Negative & 48,395 & 8,789 & 350,980 \\
        Positive top 10\% & 48,190 & 16,369 & 593,062 \\
        \hline
        Cultural positive & 10,466 & 5,064 & 87,024 \\
        Cultural negative & 1,216 & 256 & 2,360 \\
        \hline
        Animals positive & 4,826 & 827 & 71,469 \\
        Animals negative & 837 & 203 & 2,797 \\
        \hline
        Occupations positive & 2,120 & 1,250 & 13,332 \\
        Occupations negative & 145 & 83 & 322 \\
        \hline
        ConceptNet mapping & 42,407 & 14 & 436,230 \\
        \hline
    \end{tabular}
    \caption{Domain Samples of Quasimodo}
    \label{tab:domain_samples}
\end{table}

\subsection{Demo System}

We released a demo version of our paper~\cite{demo_paper}. It can be accessed at \href{https://quasimodo.r2.enst.fr}{quasimodo.r2.enst.fr}.